\documentclass[11pt]{article}

\usepackage[final]{acl}
\usepackage{times}
\usepackage{latexsym}
\usepackage[T1]{fontenc}
\usepackage[utf8]{inputenc}
\usepackage{microtype}
\usepackage{inconsolata}
\usepackage{graphicx}
\usepackage{subcaption}
\usepackage{booktabs}
\usepackage{amsmath}
\usepackage{amssymb}
\usepackage{amsthm}
\usepackage{xcolor}
\usepackage{algorithm}
\usepackage{algpseudocode}

\newcommand{\method}{IJev}

\newcommand{\anyjev}{\textsc{AnyJev-L0}}
\newcommand{\runoff}{\textsc{Runoff}}
\newcommand{\perm}{\mathrm{perm}}

\newtheorem{proposition}{Proposition}

\theoremstyle{definition}
\newtheorem{definition}{Definition}

\title{IJev: Order-Invariant Decisions
via Iterated Cyclic Reordering}

\author{
  Youzhi Zhang \\
  Hong Kong Institute of Science \& Innovation, Chinese Academy of Sciences
}

\begin{document}
\maketitle

\begin{abstract}
System One decision models answer a choice question in one forward pass, but
their answer depends on the order in which options are listed, and their
probabilities carry a prior over answer labels. An \emph{invariant ordering
operator} is any fixed function that returns the same ordered option list for
every input order of a fixed option set. Starting from any such order and
averaging a listwise readout over its $K$ cyclic shifts removes the input
order, but the decision still depends on the chosen \emph{cyclic class}: those
$K$ shifts are only one of the $(K-1)!$ classes, and different invariant
operators can select different classes. We propose \method{}, which repeatedly
reorders the options by the cyclic-averaged, label-prior-corrected distribution
and runs another cyclic average. Every iterate is exactly input-order-invariant.
When the iteration does not converge, we average the round distributions rather
than return the last round. On five benchmarks and five models, \method{} has
flip rate zero in every cell. The default uniform average over $T{=}20$ rounds
substantially improves calibration over a single cyclic average, and its
accuracy is comparable. A descriptive choice of horizon and averaging rule is
best or tied-best for accuracy in all $25$ settings.
\end{abstract}

\section{Introduction}
\label{sec:intro}

System One decision models turn a state and a typed question---yes/no, a choice
among options, or an ordered level---into a probability distribution over
answers without generating text \citep{typesafe2026jev,almeida2026jev}. Open
implementations include purpose-trained decision models and readout heads on
frozen language models
\citep{llm2jev2026,semif2026,laya2026,anyjev2026}.
However, two biases limit the reliability of these decisions. First, for a choice
question the answer depends on the order in which the options are listed, and
the dependence grows with the number of options
\citep{pezeshkpour-hruschka-2024-large,zheng2024large,anyjev2026,runoff2026}. In an agent, the order is typically inherited from a tool registry or a page layout; a change
there can silently change the prediction, and whoever sets the order can steer
it. Second, the next-token probabilities carry a
prior over answer labels: some option letters are favored independently of their
content, which distorts both the answer and its confidence
\citep{zheng2024large,zhao2021calibrate,zhou2024batch,anyjev2026}.

AnyJev's L0 readout \citep{anyjev2026} addresses both problems. It reads the
options in all $K$ cyclic shifts, so every option sits at every position once,
and divides out a label prior estimated without labels. Two limitations remain.
First, L0 is not input-order-invariant: the $K$ listings it averages are the
rotations of the \emph{input} order, so that order decides which $K$ of the
$K!$ listings enter the average. Second, those $K$ listings form a single
cyclic class. Classes that are never read can still change the decision.

\paragraph{Invariant ordering.}
The input-order dependence can be removed before any model call by an
\emph{invariant ordering operator}: a fixed function that maps an option set,
a state, and a question to an ordered list of the same options, and that
returns the same list for every input order of that set. Sorting by option text
is one such operator. Sorting by pointwise yes/no scores is another: each option
is scored alone, so the score and the induced order do not depend on the other
options or on their order. Given any invariant order $\tau$, averaging the
listwise readout over the $K$ cyclic shifts of $\tau$ is input-order-invariant,
because the input order has already been discarded.

\paragraph{Residual cyclic-class bias.}
Input-order invariance is not the end of the story. The $K!$ orders of
$K$ options partition into $(K-1)!$ \emph{cyclic classes}, where a class is the
set of cyclic shifts of one order. Cyclic averaging uses one class. Different
invariant operators---sorting by text, sorting by pointwise scores, or sorting by
a previous decision---can select different classes. A model that is not
symmetric under rotation can return a different distribution on each class.
That residual position bias is what one cyclic average leaves in place. Full
permutation marginalization would average every class, but it costs $K!$
listings.

\begin{figure*}[t]
\centering
\includegraphics[width=\textwidth]{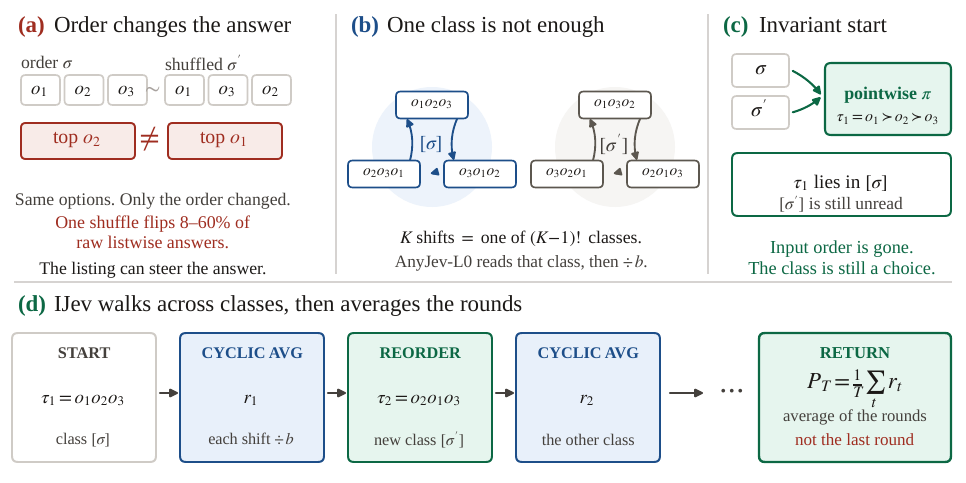}
\caption{The argument, drawn in full for $K{=}3$, where there are exactly two cyclic classes.
(a)~The same options in two orders can yield different top choices.
On the five benchmarks, one shuffle flips a raw listwise answer on between
7.9\% and 59.7\% of items.
(b)~The $K$ cyclic shifts of an order stay inside one class, and there are
$(K{-}1)!$ classes.
AnyJev-L0 averages the class of the input order and divides out a position
prior, so a shuffle still selects a different class.
(c)~Pointwise yes/no scores induce one order $\tau_1$ for every input order.
That removes the input order and leaves the class.
(d)~\method{} reorders by the prior-corrected cyclic average.
For this $\tau_1$, the next order lies in the other class.
The returned distribution is the uniform average of the round distributions,
not the last round.
Because $\tau_1$ ignores the input order, the walk and $P_T$ do too, so the flip rate is zero.}
\label{fig:story}
\end{figure*}

\paragraph{This paper.}
We propose \method{} (iterative invariant Jev), which starts from an invariant
order, applies AnyJev's prior-corrected cyclic average, and reorders the options
by the resulting distribution. The reorder step never reads the input order, so every iterate is input-order-invariant.
The iteration may reach a fixed order, enter a cycle, or still be moving when
the round budget ends. We therefore average the round distributions---uniformly,
in log space, or with exponential weights---instead of returning the last one.
That average is an ensemble over the cyclic classes the iteration visits.
Figure~\ref{fig:story} draws the argument for $K{=}3$, the case in which both
cyclic classes fit on the page.

\paragraph{Contributions.}
\begin{itemize}
\item A definition of invariant ordering operators and cyclic classes for
System One choice decisions, with text sorting and pointwise scoring as
examples. The definition separates input-order invariance from the residual
dependence on the chosen cyclic class.
\item \method{}, which iterates reordering and prior-corrected cyclic averaging.
Every iterate and every round average is exactly input-order-invariant, and on
a finite set of orders the iteration reaches a fixed point or a cycle.
\item Experiments on five benchmarks and five models. \method{} has flip rate
$0$ in every cell. Uniform averaging over $T{=}20$ rounds improves calibration
over AnyJev-L0 in nearly every setting, while accuracy stays comparable. A
post-hoc choice of horizon and rule is best or tied-best for accuracy in all
$25$ model--benchmark cells; that row is a descriptive reference, not a
confirmatory result.
\end{itemize}


\section{Related Work}
\label{sec:related}

One forward pass of a System One model returns a distribution over the answer
space, whether the question is yes/no, a choice, or a graded score
\citep{almeida2026jev,typesafe2026jev}. Public systems either fit a dedicated
decision model \citep{kev2026,laya2026,nanojev2026,yu2026visualjev} or attach a
readout to a frozen language model \citep{semif2026}. The readout may score
options independently \citep{llm2jev2026} or decode the labels of a single
listing \citep{tictop2026llm2jev,anyjev2026}.

Placing the same options in a different order changes these predictions
\citep{pezeshkpour-hruschka-2024-large,wei-etal-2024-unveiling,zong2024fool,anyjev2026,runoff2026}.
A substantial part of the effect is a learned preference for particular answer
labels \citep{zheng2024large,anyjev2026}. Such priors are usually removed by
dividing out a content-free or contextual estimate
\citep{zhao2021calibrate,zhou2024batch}. AnyJev estimates its prior on the batch
being decided, or on empty probe states, and does not use labels. We keep that
estimator fixed and change only the orders that enter the average.

Enumerating every permutation removes order dependence but requires $K!$
readings \citep{guda-etal-2025-quantifying}. Cheaper alternatives, including
$K$ cyclic shifts
\citep{izacard2023atlas,zheng2024large,anyjev2026,runoff2026}, remove only the
part of the bias that lives inside one cyclic class. The concurrent method
\runoff{} \citep{runoff2026} combines pointwise scoring with cyclic comparison:
it canonicalizes a candidate list once, with pointwise decisions, and then
compares those candidates in the cyclic listings of that one order. The
construction is input-order-invariant, and it is cheaper when the candidate set
is small, but it tends to lose accuracy and its guarantee covers only one
cyclic class.
\method{} does not prune. It iterates the canonical order and averages the
distributions of the classes it visits, trading additional reads for accuracy
and calibration. Pruning is orthogonal and can be composed with the iteration
when a smaller pool is required.

\section{Preliminaries}
\label{sec:prelim}

Consider a choice question $q$ posed about a state $x$. The candidate answers
form a set $\mathcal{O}=\{o_1,\dots,o_K\}$ containing exactly one correct
option. An order $\sigma$ is a list of all options in $\mathcal{O}$. The listwise
readout $g(x,q,\lambda)\in\Delta(\mathcal{O})$ renormalizes the model's
next-token probabilities of the letters in listing $\lambda$. Nothing is
generated or parsed. We write $g_\lambda(o)$ for the probability of option $o$.

\paragraph{AnyJev.}
For a choice question, AnyJev-L0 \citep{anyjev2026} averages the listwise
readout over the $K$ cyclic shifts of the input order and divides out a
position prior inside each listing. Let $\sigma_s$ be the $s$-th cyclic shift
of $\sigma$, and let $\mathrm{pos}_s(o)$ be the position of option $o$ in
$\sigma_s$. Then
\begin{equation}\small
p^{\mathrm{L0}}_{\sigma}(o)\!\!\;\propto\;\!\!
\exp\!\left(\frac{1}{K}\sum_{s=1}^{K}
\log\frac{g_{\sigma_s}(o)}{b_s(\mathrm{pos}_s(o))^{\beta}}\right),
\label{eq:l0}
\end{equation}
where $b_s$ is a distribution over the $K$ positions in shift $s$. 
The prior is not a distribution over option identities. It can be estimated from
the batch being decided---the mean listwise distribution over positions at the
same shift---or from content-free probe states with the same question and
listing. Neither estimate uses answer labels. 

\paragraph{Invariant ordering operators.}
\begin{definition}[Invariant ordering operator]
\label{def:invariant}
Let $\mathcal{O}$ be a finite option set and let $\Sigma(\mathcal{O})$ be the
set of all orders of $\mathcal{O}$. A map
$\Phi(x,q,\mathcal{O},\sigma)\in\Sigma(\mathcal{O})$ is \emph{invariant} if
$\Phi(x,q,\mathcal{O},\sigma)=\Phi(x,q,\mathcal{O},\sigma')$ as ordered
sequences of option identities for all $\sigma,\sigma'\in\Sigma(\mathcal{O})$.
\end{definition}
Two examples are used below. \emph{Text sort} orders options lexicographically
by their text. \emph{Pointwise sort} scores each option alone, so the score does
not depend on the remaining options or on their order. One such score is the
yes/no log-odds
\begin{align} 
f(x,q,o)\!&=\!\log P(\text{yes}\!\mid\! x,q,o)\!-\!\log P(\text{no}\!\mid\! x,q,o),\nonumber\\
\pi(o)\!&=\!\frac{e^{f(x,q,o)}}{\sum_{o'\in\mathcal{O}}e^{f(x,q,o')}},
\label{eq:pi}
\end{align}
and $\mathrm{sort}_\downarrow(\pi)$ denotes sorting by decreasing $\pi$ with
ties broken by option text. Both operators return the same ordered list under
every input order.

\begin{algorithm}[t]
\caption{Iterative Invariant Jev (\method{})}
\label{alg:method}
\footnotesize
\begin{algorithmic}[1]
\Require states $X=\{x_i\}_{i=1}^N$, question $q$, options $\mathcal{O}$
($K=|\mathcal{O}|$), round budget $T$, strength $\beta$, prior type
\Ensure round-averaged distribution $P^{(i)}$
\For{$i\gets1,\dots,N$}
  \State $f_i(o)\gets\log P(\text{yes}\mid x_i,q,o)-\log P(\text{no}\mid x_i,q,o)$
  \State $\pi_i\gets\mathrm{softmax}_{o\in\mathcal{O}}f_i(o)$
  \State $\tau_1^{(i)}\gets\mathrm{sort}_\downarrow(\pi_i)$, ties by option text
\EndFor
\For{$t\gets1,\dots,T$}
  \For{$i\gets1,\dots,N$}
    \For{$s\gets1,\dots,K$}
      \State $\lambda_{t,s}^{(i)}\gets$ $s$-th cyclic shift of $\tau_t^{(i)}$
      \State read $g_{\lambda_{t,s}^{(i)}}(o)$ for all $o\in\mathcal{O}$
    \EndFor
  \EndFor
  \State $b_{t,s}(j)\gets$ label-free prior on position $j$ at shift $s$
  \For{$i\gets1,\dots,N$}
    \State $\ell_t^{(i)}\gets$ log-average in Eq.~\eqref{eq:round}
    \State $r_t^{(i)}\gets\mathrm{softmax}(\ell_t^{(i)})$
    \State $\tau_{t+1}^{(i)}\gets\mathrm{sort}_\downarrow(r_t^{(i)})$, ties by option text
  \EndFor
  \If{$\tau_{t+1}^{(i)}=\tau_t^{(i)}$ for every $i$}
    \State \textbf{break} \Comment{fixed point; $r_t$ is stationary}
  \EndIf
\EndFor
\State $T_{\mathrm{run}}\gets$ number of executed rounds
\State $P^{(i)}\gets\frac{1}{T_{\mathrm{run}}}\sum_{t=1}^{T_{\mathrm{run}}}r_t^{(i)}$
\State \Return $P^{(i)}$ and $\{\tau_t^{(i)},r_t^{(i)}\}_{t\le T_{\mathrm{run}}}$
\end{algorithmic}
\end{algorithm}

\section{Iterative Invariant Jev}
\label{sec:method}

\method{} improves a System One choice decision by iterating a step that is
already order-invariant. Pointwise sort supplies the initial order. Each round
replaces that order by the ranking of a prior-corrected cyclic average, then
repeats. If the orders reach a fixed point we stop; otherwise we run a fixed
budget of rounds. The value we return is an average of the round distributions,
not the last round. Algorithm~\ref{alg:method} is the shared-pool procedure.
When each item has its own option set, the same updates apply per item and the
prior is estimated inside groups that share a question group, an option count,
and a shift (Section~\ref{sec:setup}).

\paragraph{Cyclic classes.}
For an order $\tau=(t_1,\dots,t_K)$, the cyclic class
$[\tau]=\{\tau_s:s=1,\dots,K\}$ is the set of its cyclic shifts.
The $K!$ orders partition into $(K-1)!$ classes, and a cyclic average over
$[\tau]$ uses exactly one of them. If $\Phi$ is an invariant ordering operator,
$[\Phi(x,q,\mathcal{O})]$ does not depend on the input order, so the cyclic
average is input-order-invariant. It still depends on which operator was used.
Text sort and pointwise sort generally select different classes, and a model
that is not rotation-symmetric can assign them different distributions. Iteration
is a way to move off that arbitrary initial class.

\paragraph{Initial order.}
The default operator is pointwise sort. Compute $\pi$ from Eq.~\eqref{eq:pi} and
set $\tau_1=\mathrm{sort}_\downarrow(\pi)$, with ties broken by option text,
in the initialization of Algorithm~\ref{alg:method}. Text sort is a zero-cost alternative.
The invariance argument applies to any operator in Definition~\ref{def:invariant};
only the class it selects changes.

\paragraph{One cyclic round.}
At round $t$, let $\lambda_{t,s}$ be the $s$-th cyclic shift of
$\tau_t=(t_1,\dots,t_K)$, and let $\mathrm{pos}_{t,s}(o)$ be the position of
option $o$ in that listing. Estimate a position prior $b_{t,s}(j)$ without
labels: the batch prior is the mean listwise position distribution of the
evaluation batch at shift $s$, and a content-free prior is the same mean on
probe states. Correct each listing and average in log space,
\begin{equation}\small
r_t(o)\;\propto\;\exp\!\left(\frac{1}{K}\sum_{s=1}^{K}
\log\frac{g_{\lambda_{t,s}}(o)}{b_{t,s}(\mathrm{pos}_{t,s}(o))^{\beta}}\right).
\label{eq:round}
\end{equation}
This is AnyJev-L0, Eq.~\eqref{eq:l0}, applied to the current invariant order
rather than to the input order. We use $\beta=1.0$ and the batch prior.
Here $g_\lambda(o)$ is the renormalized next-token probability of the letter in
front of $o$ when the model is shown listing $\lambda$.

\paragraph{Reordering.}
The next invariant order ranks the just-computed distribution,
\begin{equation}\small
\tau_{t+1}=\mathrm{sort}_\downarrow(r_t),
\label{eq:reorder}
\end{equation}
again with ties broken by option text. Round $t+1$ averages the cyclic shifts
of $\tau_{t+1}$. Because $r_t$ does not depend on the input order, neither does
$\tau_{t+1}$.

\paragraph{Round averaging.}
A fixed point makes $r_t$ stationary, and the last round is then enough. A cycle
repeats a finite set of distributions. A trajectory that is still moving makes
the last round an arbitrary sample. We therefore return an average over the
rounds that were run. Writing $T$ for that number, the default is the uniform
average
\begin{equation}\small
P_T(o)=\frac{1}{T}\sum_{t=1}^{T}r_t(o).
\label{eq:uniform}
\end{equation}
We also evaluate the log-average
\begin{equation}\small
P_T^{\log}(o)\;\propto\;
\exp\!\left(\frac{1}{T}\sum_{t=1}^{T}\log r_t(o)\right),
\label{eq:logavg}
\end{equation}
using the same probability floor as the prior correction, and the exponential
average $P_T(o)\propto\sum_{t=1}^{T}w_t r_t(o)$ with
$w_t\propto e^{-(T-t)/\tau_w}$. 

\begin{proposition}[Invariance of every iterate and of the round average]
\label{prop:invariance}
For a deterministic model, if the initial order $\tau_1$ is produced by an
invariant ordering operator, then every $\tau_t$, every $r_t$, and every round
average $P_T$ is independent of the input order of $\mathcal{O}$. (Proofs are in Appendix \ref{app:proofs}.)
\end{proposition}


\begin{proposition}[Termination or cycle]
\label{prop:termination}
With a fixed prior, the map $\tau_t\mapsto\tau_{t+1}$ is deterministic on the
$K!$ orders of one state, so the sequence reaches a fixed point or a cycle
within $K!+1$ rounds. With the batch prior, the prior at round $t$ depends on
all $N$ orders, the state is the joint order vector, and the same counting
argument gives a cycle within $(K!)^N+1$ rounds.
\end{proposition}


\paragraph{What iteration changes.}
One cyclic average depends on the class $[\tau_1]$. Reordering by $r_t$ and
averaging again moves to another class, or returns to the same one. The round
average is an ensemble over the classes that were visited. It does not cover all
$(K-1)!$ classes, so it is not full permutation marginalization. It does remove
the privilege of whichever class the initial operator happened to select. In our
runs the trajectory is still moving at $T=20$, which is why the last round is
not a stable rule.

\paragraph{Cost.}
Pointwise sort costs $K$ yes/no reads. Each round costs $K$ listwise reads and
$K^2$ option slots, so $T$ rounds cost $TK$ listwise reads. For $K=20$ and
$T=20$ that is 400 listwise reads, twenty times the 20 reads of L0. The batch
prior is computed from those reads and adds none. A content-free prior would add
probe reads. \method{} spends this compute on invariance and calibration; it is
not a cheaper readout. Future work could develop cheaper methods   while maintaining the accuracy.  

\begin{table*}[t]
\centering
\footnotesize
\setlength{\tabcolsep}{2.5pt}
\renewcommand{\arraystretch}{0.88}
\begin{tabular}{llcccccccccccc}
\toprule
 & & \multicolumn{2}{c}{Banking20} & \multicolumn{2}{c}{Newsgroups} & \multicolumn{2}{c}{MMLU-Pro} & \multicolumn{2}{c}{Typed Dec.} & \multicolumn{2}{c}{JevBench} & \multicolumn{2}{c}{Avg} \\
\cmidrule(lr){3-4} \cmidrule(lr){5-6} \cmidrule(lr){7-8} \cmidrule(lr){9-10} \cmidrule(lr){11-12} \cmidrule(lr){13-14}
Model & Method & Acc$\uparrow$ & Flip$\downarrow$ & Acc$\uparrow$ & Flip$\downarrow$ & Acc$\uparrow$ & Flip$\downarrow$ & Acc$\uparrow$ & Flip$\downarrow$ & Acc$\uparrow$ & Flip$\downarrow$ & Acc$\uparrow$ & Flip$\downarrow$ \\
\midrule
Qwen3-1.7B & Listwise (raw) & 62.1 & 33.7 & 59.0 & 28.7 & 27.6 & 59.7 & 47.3 & 39.0 & 54.7 & 28.1 & 50.1 & 37.8 \\
 & AnyJev-L0 & 69.5 & 11.8 & 61.3 & 19.7 & 33.2 & 31.2 & 49.0 & 25.7 & 59.0 & 9.4 & 54.4 & 19.5 \\
 & \method{} (uniform, $T{=}20$) & 70.3 & \textbf{0.0} & 62.0 & \textbf{0.0} & 33.2 & \textbf{0.0} & \textbf{57.7} & \textbf{0.0} & 57.6 & \textbf{0.0} & 56.2 & \textbf{0.0} \\
 & \method{} (log-avg, $T{=}20$) & 70.5 & \textbf{0.0} & 63.0 & \textbf{0.0} & 33.3 & \textbf{0.0} & 57.5 & \textbf{0.0} & 58.3 & \textbf{0.0} & 56.5 & \textbf{0.0} \\
 & \method{} (best) & \textbf{72.0} & \textbf{0.0} & \textbf{65.3} & \textbf{0.0} & \textbf{34.6} & \textbf{0.0} & \textbf{57.7} & \textbf{0.0} & \textbf{61.9} & \textbf{0.0} & \textbf{58.3} & \textbf{0.0} \\
\midrule
Qwen3-4B & Listwise (raw) & 73.1 & 21.9 & 63.7 & 20.0 & 40.9 & 48.6 & 55.8 & 24.7 & 65.5 & 12.9 & 59.8 & 25.6 \\
 & AnyJev-L0 & 76.3 & 13.0 & 66.7 & 11.7 & 46.4 & 23.9 & \textbf{59.8} & 15.7 & 64.7 & 5.0 & 62.8 & 13.8 \\
 & \method{} (uniform, $T{=}20$) & 77.1 & \textbf{0.0} & 67.3 & \textbf{0.0} & 46.0 & \textbf{0.0} & 59.3 & \textbf{0.0} & 66.2 & \textbf{0.0} & 63.2 & \textbf{0.0} \\
 & \method{} (log-avg, $T{=}20$) & 77.2 & \textbf{0.0} & 67.7 & \textbf{0.0} & 46.0 & \textbf{0.0} & 59.2 & \textbf{0.0} & 64.7 & \textbf{0.0} & 63.0 & \textbf{0.0} \\
 & \method{} (best) & \textbf{77.6} & \textbf{0.0} & \textbf{69.7} & \textbf{0.0} & \textbf{46.9} & \textbf{0.0} & \textbf{59.8} & \textbf{0.0} & \textbf{69.8} & \textbf{0.0} & \textbf{64.7} & \textbf{0.0} \\
\midrule
Qwen3-8B & Listwise (raw) & 75.8 & 18.4 & 64.3 & 20.7 & 44.5 & 46.5 & 60.2 & 17.0 & 71.2 & 7.9 & 63.2 & 22.1 \\
 & AnyJev-L0 & 81.2 & 7.6 & 66.3 & 12.7 & 50.4 & 23.7 & 65.2 & 9.5 & 71.9 & 1.4 & 67.0 & 11.0 \\
 & \method{} (uniform, $T{=}20$) & 81.7 & \textbf{0.0} & 66.0 & \textbf{0.0} & 51.9 & \textbf{0.0} & 65.0 & \textbf{0.0} & 70.5 & \textbf{0.0} & 67.0 & \textbf{0.0} \\
 & \method{} (log-avg, $T{=}20$) & 81.9 & \textbf{0.0} & 65.3 & \textbf{0.0} & \textbf{52.3} & \textbf{0.0} & 65.8 & \textbf{0.0} & 71.9 & \textbf{0.0} & 67.4 & \textbf{0.0} \\
 & \method{} (best) & \textbf{82.0} & \textbf{0.0} & \textbf{66.7} & \textbf{0.0} & \textbf{52.3} & \textbf{0.0} & \textbf{66.2} & \textbf{0.0} & \textbf{73.4} & \textbf{0.0} & \textbf{68.1} & \textbf{0.0} \\
\midrule
Mistral-7B & Listwise (raw) & 72.6 & 31.9 & 64.7 & 30.0 & 30.6 & 55.3 & 56.0 & 22.2 & 59.0 & 12.9 & 56.6 & 30.5 \\
 & AnyJev-L0 & 76.3 & 13.6 & 67.0 & 11.3 & 32.8 & 39.4 & 58.3 & 21.3 & 59.0 & 5.8 & 58.7 & 18.3 \\
 & \method{} (uniform, $T{=}20$) & 76.8 & \textbf{0.0} & 69.0 & \textbf{0.0} & 36.6 & \textbf{0.0} & 59.7 & \textbf{0.0} & 60.4 & \textbf{0.0} & 60.5 & \textbf{0.0} \\
 & \method{} (log-avg, $T{=}20$) & 77.5 & \textbf{0.0} & 67.7 & \textbf{0.0} & 37.1 & \textbf{0.0} & 59.5 & \textbf{0.0} & 59.7 & \textbf{0.0} & 60.3 & \textbf{0.0} \\
 & \method{} (best) & \textbf{78.0} & \textbf{0.0} & \textbf{70.0} & \textbf{0.0} & \textbf{37.6} & \textbf{0.0} & \textbf{60.3} & \textbf{0.0} & \textbf{62.6} & \textbf{0.0} & \textbf{61.7} & \textbf{0.0} \\
\midrule
Phi-3.5-mini & Listwise (raw) & 76.8 & 26.2 & 64.3 & 28.3 & 40.4 & 47.1 & 56.2 & 16.2 & 58.3 & 12.2 & 59.2 & 26.0 \\
 & AnyJev-L0 & 80.2 & 9.6 & 64.3 & 11.3 & 43.3 & 27.4 & 58.8 & 16.0 & 58.3 & 5.0 & 61.0 & 13.9 \\
 & \method{} (uniform, $T{=}20$) & 82.0 & \textbf{0.0} & 66.0 & \textbf{0.0} & 45.1 & \textbf{0.0} & 59.2 & \textbf{0.0} & 57.6 & \textbf{0.0} & 62.0 & \textbf{0.0} \\
 & \method{} (log-avg, $T{=}20$) & 81.7 & \textbf{0.0} & 66.7 & \textbf{0.0} & 45.3 & \textbf{0.0} & 58.8 & \textbf{0.0} & 56.8 & \textbf{0.0} & 61.9 & \textbf{0.0} \\
 & \method{} (best) & \textbf{82.5} & \textbf{0.0} & \textbf{68.3} & \textbf{0.0} & \textbf{45.7} & \textbf{0.0} & \textbf{59.5} & \textbf{0.0} & \textbf{62.6} & \textbf{0.0} & \textbf{63.7} & \textbf{0.0} \\
\bottomrule
\end{tabular}
\caption{Accuracy and flip rate on five decision benchmarks. 
Avg is the unweighted mean over the five datasets. Bold marks the best value for each model and column.
}
\label{tab:all}
\end{table*}

\section{Experiments}
\label{sec:experiments}

\subsection{Setup}
\label{sec:setup}

\paragraph{Benchmarks.}
We evaluate five single-label choice benchmarks. \textbf{Banking20} is the
BANKING77 test split restricted to the 20 most frequent intents: 799 items with
a shared set of 20 intent options, all evaluated in the seed-0 order.
\textbf{Newsgroups} is the standard 300-item test sample of the 20-way topic
classification task, again with a shared option set. \textbf{MMLU-Pro}
\citep{wang2024mmlupro} contributes 1,400 multiple-choice questions, 100 per
category, with 3--10 options each. \textbf{Typed Decisions}
\citep{typeddecisions2026} contains 600 choice questions with 4--5 options, and
the public items of \textbf{JevBench} \citep{jevbench2026} contain 139 choice
questions with 3--6 options. The last three assign every question its own
option set and option count; there we use the per-item form of the operator,
with cyclic shifts of each question's own order and a prior grouped by
question group, option count, and shift. There is no training and no
calibration set; every prior is estimated without labels.

\paragraph{Models.}
We use five instruction-tuned models from three families, spanning 1.7B--8B
parameters: Qwen3-1.7B, Qwen3-4B and Qwen3-8B \citep{yang2025qwen3};
Mistral-7B-Instruct-v0.3 \citep{jiang2023mistral}; and
Phi-3.5-mini-instruct \citep{abdin2024phi}.
All runs are fp16 on a single NVIDIA RTX 4090 D (24 GB), with PyTorch 2.6.0 and
CUDA 12.4, and use the model's chat template.

\paragraph{Implementation.}
The listwise readout $g_\lambda$ reads the next-token probabilities of the
option letters and renormalizes them. The pointwise probe reads the yes/no
next-token log-odds for one candidate at a time. Every prompt is rendered with
the chat template, sorted by prompt text within a batch, and read with a single
forward pass, so the batch composition does not depend on the option order.
Decoding is deterministic on a fixed model and prompt set.
Appendix~\ref{app:implementation} gives the prompt, prior, and precision settings.

\begin{table*}[t]
\centering
\footnotesize
\setlength{\tabcolsep}{0.7pt}
\renewcommand{\arraystretch}{0.7}
\begin{tabular}{llccccccccccccccc}
\toprule
 & & \multicolumn{3}{c}{Banking20} & \multicolumn{3}{c}{Newsgroups} & \multicolumn{3}{c}{MMLU-Pro} & \multicolumn{3}{c}{Typed Dec.} & \multicolumn{3}{c}{JevBench} \\
\cmidrule(lr){3-5} \cmidrule(lr){6-8} \cmidrule(lr){9-11} \cmidrule(lr){12-14} \cmidrule(lr){15-17}
Model & Method & ECE$\downarrow$ & Brier$\downarrow$ & NLL$\downarrow$ & ECE$\downarrow$ & Brier$\downarrow$ & NLL$\downarrow$ & ECE$\downarrow$ & Brier$\downarrow$ & NLL$\downarrow$ & ECE$\downarrow$ & Brier$\downarrow$ & NLL$\downarrow$ & ECE$\downarrow$ & Brier$\downarrow$ & NLL$\downarrow$ \\
\midrule
Qwen3-1.7B & Listwise (raw) & 0.36 & 0.73 & 6.56 & 0.39 & 0.79 & 7.03 & 0.63 & 1.31 & 7.59 & 0.51 & 1.02 & 7.29 & 0.40 & 0.81 & 4.14 \\
 & AnyJev-L0 & 0.27 & 0.57 & 3.38 & 0.35 & 0.72 & 4.72 & 0.46 & 1.08 & 4.70 & 0.37 & 0.83 & 3.03 & 0.36 & 0.76 & 3.71 \\
 & \method{} (uniform) & \textbf{0.19} & \textbf{0.47} & \textbf{1.84} & \textbf{0.28} & \textbf{0.64} & \textbf{2.98} & \textbf{0.31} & \textbf{0.94} & \textbf{3.32} & \textbf{0.25} & \textbf{0.72} & \textbf{2.50} & \textbf{0.28} & \textbf{0.64} & \textbf{2.13} \\
 & \method{} (log-avg) & 0.25 & 0.53 & 2.62 & 0.32 & 0.69 & 4.12 & 0.37 & 1.00 & 3.77 & 0.34 & 0.75 & 2.86 & 0.33 & 0.72 & 2.63 \\
 & \method{} (best) & \textbf{0.19} & 0.49 & 2.28 & 0.31 & 0.66 & 4.11 & 0.33 & 0.97 & 3.60 & \textbf{0.25} & \textbf{0.72} & \textbf{2.50} & 0.31 & 0.69 & 3.05 \\
\addlinespace[3pt]
Qwen3-4B & Listwise (raw) & 0.26 & 0.52 & 5.07 & 0.34 & 0.69 & 6.17 & 0.50 & 1.06 & 6.80 & 0.40 & 0.83 & 5.96 & 0.32 & 0.64 & 4.01 \\
 & AnyJev-L0 & 0.21 & 0.45 & 2.99 & 0.31 & 0.63 & 4.72 & 0.40 & 0.91 & 4.58 & 0.29 & \textbf{0.67} & 2.47 & 0.32 & 0.64 & 3.58 \\
 & \method{} (uniform) & \textbf{0.16} & \textbf{0.41} & \textbf{2.06} & \textbf{0.24} & \textbf{0.55} & \textbf{2.70} & \textbf{0.29} & \textbf{0.83} & \textbf{3.39} & \textbf{0.28} & \textbf{0.67} & \textbf{2.29} & 0.23 & \textbf{0.52} & \textbf{2.33} \\
 & \method{} (log-avg) & 0.20 & 0.43 & 2.88 & 0.30 & 0.61 & 3.77 & 0.35 & 0.88 & 3.89 & 0.35 & 0.74 & 2.89 & 0.29 & 0.58 & 2.69 \\
 & \method{} (best) & \textbf{0.16} & \textbf{0.41} & 2.16 & 0.28 & 0.59 & 4.14 & 0.35 & 0.87 & 3.89 & 0.29 & 0.68 & 2.31 & \textbf{0.22} & \textbf{0.52} & 2.43 \\
\addlinespace[3pt]
Qwen3-8B & Listwise (raw) & 0.23 & 0.47 & 4.30 & 0.33 & 0.68 & 5.72 & 0.47 & 0.99 & 5.89 & 0.37 & 0.76 & 4.93 & 0.27 & 0.55 & 3.61 \\
 & AnyJev-L0 & 0.18 & 0.36 & 2.48 & 0.31 & 0.63 & 4.27 & 0.37 & 0.84 & 3.98 & 0.27 & 0.61 & 2.82 & 0.25 & 0.53 & 3.10 \\
 & \method{} (uniform) & 0.14 & \textbf{0.33} & \textbf{1.69} & 0.26 & \textbf{0.58} & \textbf{2.92} & \textbf{0.24} & \textbf{0.74} & \textbf{2.77} & \textbf{0.23} & \textbf{0.59} & \textbf{2.68} & \textbf{0.20} & \textbf{0.49} & \textbf{2.04} \\
 & \method{} (log-avg) & 0.17 & 0.35 & 2.43 & 0.32 & 0.64 & 4.09 & 0.31 & 0.78 & 3.25 & 0.28 & 0.62 & 3.04 & 0.23 & 0.50 & 2.28 \\
 & \method{} (best) & \textbf{0.13} & \textbf{0.33} & 1.70 & \textbf{0.25} & \textbf{0.58} & 3.10 & 0.31 & 0.78 & 3.25 & 0.28 & 0.62 & 3.04 & \textbf{0.20} & \textbf{0.49} & 2.09 \\
\addlinespace[3pt]
Mistral-7B & Listwise (raw) & 0.25 & 0.52 & 3.49 & 0.32 & 0.66 & 4.96 & 0.54 & 1.20 & 6.60 & 0.35 & 0.77 & 3.76 & 0.33 & 0.71 & 3.35 \\
 & AnyJev-L0 & 0.20 & 0.43 & 1.95 & 0.29 & 0.61 & 3.70 & 0.48 & 1.09 & 4.54 & 0.25 & 0.68 & 1.87 & 0.34 & 0.72 & 3.12 \\
 & \method{} (uniform) & \textbf{0.11} & \textbf{0.36} & \textbf{1.10} & \textbf{0.19} & \textbf{0.51} & \textbf{1.95} & \textbf{0.28} & \textbf{0.91} & \textbf{3.05} & \textbf{0.23} & \textbf{0.64} & \textbf{1.62} & \textbf{0.26} & \textbf{0.64} & \textbf{2.09} \\
 & \method{} (log-avg) & 0.18 & 0.40 & 1.53 & 0.27 & 0.58 & 2.61 & 0.36 & 0.98 & 3.51 & 0.28 & 0.67 & 1.85 & 0.29 & 0.67 & 2.32 \\
 & \method{} (best) & \textbf{0.11} & 0.37 & 1.26 & \textbf{0.19} & 0.52 & 2.33 & 0.35 & 0.98 & 3.52 & \textbf{0.23} & 0.65 & 1.69 & 0.29 & 0.65 & 2.41 \\
\addlinespace[3pt]
Phi-3.5-mini & Listwise (raw) & 0.19 & 0.41 & 2.70 & 0.31 & 0.66 & 4.67 & 0.43 & 0.99 & 4.47 & 0.36 & 0.76 & 3.33 & 0.36 & 0.77 & 4.75 \\
 & AnyJev-L0 & 0.16 & 0.34 & 1.59 & 0.33 & 0.67 & 3.70 & 0.35 & 0.88 & 3.39 & 0.25 & \textbf{0.65} & \textbf{1.71} & 0.37 & 0.76 & 4.25 \\
 & \method{} (uniform) & \textbf{0.09} & \textbf{0.30} & \textbf{1.04} & \textbf{0.20} & \textbf{0.52} & \textbf{1.75} & \textbf{0.20} & \textbf{0.78} & \textbf{2.44} & \textbf{0.24} & 0.67 & 1.87 & \textbf{0.30} & \textbf{0.67} & \textbf{3.04} \\
 & \method{} (log-avg) & 0.13 & 0.31 & 1.41 & 0.26 & 0.59 & 2.57 & 0.25 & 0.81 & 2.69 & 0.32 & 0.71 & 2.25 & 0.35 & 0.72 & 3.39 \\
 & \method{} (best) & 0.12 & 0.31 & 1.43 & \textbf{0.20} & 0.53 & 2.09 & 0.21 & 0.79 & 2.51 & \textbf{0.24} & 0.67 & 1.87 & 0.32 & 0.70 & 3.32 \\
\bottomrule
\end{tabular}
\caption{ECE, Brier score, and NLL for the five methods on all five benchmarks. Lower is better in every column. 
The iterative rows are the uniform and log-average rules at $T=20$, and ``best'' selects the post-hoc accuracy-maximizing horizon. Bold marks the lowest value per model and column.}
\label{tab:metrics}
\end{table*}

\paragraph{Metrics.}
  Accuracy ($\mathrm{Acc}$) is the fraction of
items whose top-1 option is correct and measures decision quality on the
original listing. $\mathrm{Flip}$ is the fraction whose top-1 option changes under one
independent shuffle; it measures input-order sensitivity directly, and an
order-invariant readout has flip 0. $\mathrm{ECE}$ compares confidence with accuracy over
$B=15$ equal-mass bins $S_b$, so ECE is the
bin-size-weighted average calibration gap (0 is perfect calibration). $\mathrm{Brier}$ is
the mean squared error of the whole predicted distribution over the $K_i$
options, a proper score that rewards distributions that are both calibrated and
sharp. $\mathrm{NLL}$ is the mean negative log-probability of the correct option, also a
proper score; it penalizes confident mistakes most heavily and is unbounded
above. Accuracy and flip rate are percentages; lower ECE, Brier, and NLL are better. Details are in Appendix \ref{app:metrics}.

\paragraph{Methods.}
We compare the one-pass listwise readout on the input order
(\textsc{Listwise}), AnyJev-L0 with the label-free batch prior
(Eq.~\eqref{eq:l0}), and \method{} started from the pointwise order with
$\beta=1.0$. AnyJev-L0 and the iterative runs both use $\beta=1.0$
throughout, so the prior correction is identical across all five benchmarks.
The iteration has one distribution per round; we report
the last round, the uniform average, the log-average, and the exponential
averages with $\tau_w=10$ and $20$, each at $T=20$. For every model and
benchmark we also compute each rule at every horizon $T=1,\dots,20$ and define
\method{} (best) as the (rule, $T$) pair with the highest accuracy, breaking
ties by lower NLL. That row selects the horizon on the same test set, so it is
a descriptive upper reference rather than a confirmatory result; the fixed
$T=20$ uniform average is the default rule in the detailed tables.

\begin{table*}[t]
\centering
\footnotesize
\setlength{\tabcolsep}{3pt}
\renewcommand{\arraystretch}{0.88}
\begin{tabular}{lcccccccccccc}
\toprule
 & \multicolumn{6}{c}{\method{} uniform} & \multicolumn{6}{c}{\method{} log-avg} \\
\cmidrule(lr){2-7}\cmidrule(lr){8-13}
Model & $T{=}1$ & $T{=}2$ & $T{=}5$ & $T{=}10$ & $T{=}15$ & $T{=}20$ & $T{=}1$ & $T{=}2$ & $T{=}5$ & $T{=}10$ & $T{=}15$ & $T{=}20$ \\
\midrule
\multicolumn{13}{l}{\textit{Banking20}} \\
Qwen3-1.7B & 69.5 & 70.5 & 70.6 & \textbf{70.8} & 70.2 & 70.3 & 69.5 & \textbf{70.8} & 70.7 & \textbf{70.8} & 70.1 & 70.5 \\
Qwen3-4B & 76.0 & 75.8 & 76.8 & 76.2 & \textbf{77.3} & 77.1 & 76.0 & 76.2 & 77.0 & 76.6 & 77.2 & 77.2 \\
Qwen3-8B & 80.4 & 81.4 & \textbf{82.0} & 81.4 & 81.5 & 81.7 & 80.4 & 81.6 & 81.7 & 81.5 & 81.5 & 81.9 \\
Mistral-7B & 73.3 & 75.1 & 76.7 & 76.8 & 77.0 & 76.8 & 73.3 & 76.8 & 77.0 & 77.0 & \textbf{77.6} & 77.5 \\
Phi-3.5-mini & 78.2 & 81.0 & 80.9 & 81.6 & 81.6 & \textbf{82.0} & 78.2 & 81.6 & \textbf{82.0} & \textbf{82.0} & 81.6 & 81.7 \\
\midrule
\multicolumn{13}{l}{\textit{Newsgroups}} \\
Qwen3-1.7B & 65.0 & 64.3 & 63.0 & 61.3 & 61.7 & 62.0 & 65.0 & \textbf{65.3} & 63.0 & 62.7 & 62.0 & 63.0 \\
Qwen3-4B & 65.7 & \textbf{68.0} & 67.0 & 66.3 & 67.3 & 67.3 & 65.7 & 67.7 & 67.7 & 66.7 & 67.3 & 67.7 \\
Qwen3-8B & 64.7 & 65.0 & 65.7 & \textbf{66.3} & 65.7 & 66.0 & 64.7 & 65.3 & \textbf{66.3} & 66.0 & 65.7 & 65.3 \\
Mistral-7B & 66.0 & 67.3 & 68.0 & 68.0 & 68.7 & \textbf{69.0} & 66.0 & 66.7 & 68.0 & 68.0 & 68.0 & 67.7 \\
Phi-3.5-mini & 64.3 & 65.7 & 67.0 & 66.3 & 66.3 & 66.0 & 64.3 & 66.3 & 67.0 & \textbf{67.3} & 67.0 & 66.7 \\
\midrule
\multicolumn{13}{l}{\textit{MMLU-Pro}} \\
Qwen3-1.7B & 32.8 & 32.9 & 32.5 & 32.9 & 33.5 & 33.2 & 32.8 & 33.0 & 32.7 & \textbf{33.6} & 33.3 & 33.3 \\
Qwen3-4B & 45.1 & 45.6 & 46.1 & 46.4 & 46.0 & 46.0 & 45.1 & 46.1 & 46.0 & 46.4 & \textbf{46.5} & 46.0 \\
Qwen3-8B & 49.6 & 50.6 & 50.8 & 51.7 & 52.0 & 51.9 & 49.6 & 51.1 & 51.9 & 51.9 & 51.9 & \textbf{52.3} \\
Mistral-7B & 32.9 & 35.4 & 35.3 & 36.0 & 36.6 & 36.6 & 32.9 & 35.5 & 36.6 & 36.7 & \textbf{37.6} & 37.1 \\
Phi-3.5-mini & 41.6 & 43.1 & 44.0 & 45.0 & 45.1 & 45.1 & 41.6 & 43.6 & 45.0 & 44.9 & 45.2 & \textbf{45.3} \\
\midrule
\multicolumn{13}{l}{\textit{Typed Decisions}} \\
Qwen3-1.7B & 55.2 & 54.3 & 56.5 & 56.8 & 56.5 & \textbf{57.7} & 55.2 & 55.0 & 57.0 & 57.2 & 57.5 & 57.5 \\
Qwen3-4B & 59.2 & 59.5 & 59.5 & 59.5 & 59.2 & 59.3 & 59.2 & 59.2 & \textbf{59.7} & 59.2 & 59.0 & 59.2 \\
Qwen3-8B & 64.0 & 64.0 & 65.3 & 64.5 & 65.7 & 65.0 & 64.0 & 64.3 & 65.5 & 65.8 & \textbf{66.0} & 65.8 \\
Mistral-7B & 56.8 & 58.7 & 59.2 & 58.8 & 59.2 & \textbf{59.7} & 56.8 & 59.2 & \textbf{59.7} & 59.3 & 59.3 & 59.5 \\
Phi-3.5-mini & 56.2 & 58.7 & 58.5 & 59.0 & 58.5 & \textbf{59.2} & 56.2 & 58.8 & 58.8 & 59.0 & 58.8 & 58.8 \\
\midrule
\multicolumn{13}{l}{\textit{JevBench}} \\
Qwen3-1.7B & \textbf{60.4} & 56.8 & 58.3 & 56.8 & 58.3 & 57.6 & \textbf{60.4} & 58.3 & 59.0 & 58.3 & 58.3 & 58.3 \\
Qwen3-4B & \textbf{67.6} & 65.5 & 66.2 & 65.5 & 66.2 & 66.2 & \textbf{67.6} & 64.7 & 64.0 & 64.0 & 64.0 & 64.7 \\
Qwen3-8B & 69.8 & \textbf{72.7} & 71.9 & 71.2 & 71.9 & 70.5 & 69.8 & 71.9 & \textbf{72.7} & 70.5 & 71.9 & 71.9 \\
Mistral-7B & 57.6 & 59.0 & 59.7 & 59.0 & \textbf{60.4} & \textbf{60.4} & 57.6 & 59.0 & 59.7 & \textbf{60.4} & 59.7 & 59.7 \\
Phi-3.5-mini & 56.8 & \textbf{59.7} & 56.8 & 57.6 & 57.6 & 57.6 & 56.8 & 59.0 & 58.3 & 56.8 & 56.8 & 56.8 \\
\bottomrule
\end{tabular}
\caption{Accuracy (\%) of \method{} at $T=1,2,5,10,15,20$ for the uniform and log-average rules, every model and benchmark. All iterative cells are input-order-invariant (flip 0). Bold marks the best cell in each model row.}
\label{tab:rounds-grid}
\end{table*}
\begin{table*}[!htp]
\centering
\footnotesize
\setlength{\tabcolsep}{4pt}
\begin{tabular}{llcccccccc}
\toprule
Model & Task & Listwise & AnyJev-L0 & $T{=}1$ & $T{=}2$ & $T{=}5$ & $T{=}10$ & $T{=}15$ & $T{=}20$ \\
\midrule
1.7B & Banking20 & 0.008 & 0.141 & 0.211 & 0.363 & 0.819 & 1.578 & 2.337 & 3.096 \\
4B   & Banking20 & 0.017 & 0.333 & 0.483 & 0.825 & 1.850 & 3.557 & 5.265 & 6.972 \\
8B   & Banking20 & 0.030 & 0.589 & 0.821 & 1.413 & 3.193 & 6.146 & 9.096 & 12.045 \\
1.7B & Newsgroups & 0.014 & 0.244 & 0.445 & 0.697 & 1.454 & 2.715 & 3.978 & 5.239 \\
4B   & Newsgroups & 0.031 & 0.557 & 1.003 & 1.571 & 3.273 & 6.110 & 8.947 & 11.785 \\
8B   & Newsgroups & 0.049 & 0.918 & 1.617 & 2.534 & 5.284 & 9.869 & 14.455 & 19.041 \\
\bottomrule
\end{tabular}
\caption{Seconds per decision (wall time / number of states).
Iterative columns use the uniform round average. Banking20 uses 799 states;
Newsgroups uses 300.}
\label{tab:runtime}
\end{table*}
\subsection{Main results}
\label{sec:mainres}

Table~\ref{tab:all} compares the listwise readout, AnyJev-L0, and \method{} on five
models and five benchmarks. Flip rate is the clean separation. Listwise flip
runs from 7.9\% to 59.7\%, and AnyJev-L0 reduces it to 1.4--39.4\%, but never
to zero, because L0 still rotates the input order. Every \method{} cell has
flip $0.0$, as Proposition~\ref{prop:invariance} guarantees. Consistent with
that proposition, the 20-round runs were not repeated under a shuffle
(Section~\ref{sec:results}).

Calibration, not accuracy, is the reliable gain of the default rule.
Table~\ref{tab:metrics} reports ECE, Brier, and NLL for the same systems.
Relative to AnyJev-L0, the uniform average at $T=20$ lowers ECE in all 25
cells, Brier in 23, and NLL in 24. The misses are on Typed Decisions: Brier
ties AnyJev-L0 for Qwen3-4B, and both Brier and NLL are worse for
Phi-3.5-mini. The uniform rule has the lowest NLL of the five methods in 24 of
the 25 cells, and the log-average is less calibrated than the uniform average
even when the two accuracies are close.

Accuracy should be read in two layers. The confirmatory rule is the uniform
average at the pre-specified budget $T=20$. Averaged over the five benchmarks it
matches or exceeds AnyJev-L0 for every model (Table~\ref{tab:all}, Avg column):
the margins are $+1.8$, $+0.4$, $0.0$, $+1.8$, and $+1.0$ points on Qwen3-1.7B,
Qwen3-4B, Qwen3-8B, Mistral-7B, and Phi-3.5-mini. Individual cells move both
ways. The largest gain is Typed Decisions on Qwen3-1.7B ($49.0$ to $57.7$);
JevBench loses a point or two for three of the five models. The best row, which
selects the horizon and the averaging rule on the same test set, is the
highest-accuracy readout in all 25 model--benchmark cells and ties AnyJev-L0 on
Typed Decisions with Qwen3-4B. Averaged over models, that descriptive row gains
$1.7$ points over L0 on Banking20, $2.9$ on Newsgroups, $2.2$ on MMLU-Pro,
$2.5$ on Typed Decisions, and $3.5$ on JevBench. We do not treat those gains as
confirmed. Section~\ref{sec:results} reports the paired tests for the fixed
rule, which are positive for accuracy, and large for
calibration.

\subsection{Banking20: Analysis}
\label{sec:results}

Banking20 is the only benchmark with the full 799-item pool, so the paired tests
are reported here. Appendix Table~\ref{tab:main-full} lists every averaging rule
at $T=20$. The 20-round runs were executed on the original option order only. A
second 20-round pass would have doubled an already expensive sweep, and it would
have been identical: the initial order is invariant, prompts are sorted
canonically, and the cyclic average does not read the input order
(Proposition~\ref{prop:invariance}). Listwise and AnyJev-L0 flips were measured
under one per-item shuffle.

\paragraph{Accuracy.}
Under the fixed $T=20$ rule the uniform average gains 0.8, 0.8, and 0.5 points
over \anyjev{} on 1.7B, 4B, and 8B (7, 7, and 4 items). Exact McNemar tests give
29 fixes and 22 breaks ($p=0.40$), 28 and 22 ($p=0.48$), and 21 and 17
($p=0.63$): the gains are positive but not significant. The exponential
averages stay within 0.3 points of the uniform rule. The post-hoc best rows add
a further 0.3--1.7 points, but they choose the horizon on this test set. Their
McNemar tests (38/18, 30/20, and 21/15 fixes/breaks) give $p=0.01$, $0.20$, and
$0.41$ and are descriptive, not confirmatory. Only the 1.7B descriptive
comparison would clear a conventional threshold, and it was not pre-specified.

\paragraph{Calibration.}
The calibration gains are larger and statistically reliable. Paired bootstrap
intervals exclude zero on all three models. Brier improves by 0.100 (95\% CI
0.071--0.129), 0.042 (0.013--0.071), and 0.030 (0.007--0.054) on 1.7B, 4B, and
8B, and NLL by 1.543 (1.292--1.799), 0.924 (0.711--1.141), and 0.799
(0.616--0.991). ECE falls in every cell (Table~\ref{tab:metrics}). Exponential
averages stay within 0.03 NLL of the uniform rule. The log-average is
consistently worse on both Brier and NLL.



\subsection{Round-count ablation}
\label{sec:rounds}

Table~\ref{tab:rounds-grid} reports accuracy of the uniform and log-average
rules at $T=1,2,5,10,15,20$ for every model and benchmark.
Appendix Table~\ref{tab:rounds-bank} adds ECE and NLL on Banking20, and
Appendix~\ref{app:rounds-full} records every horizon.

Calibration moves first. Most of the NLL reduction is present by five to ten
rounds, and for the uniform average the best NLL horizon is $T=20$ in every
setting we checked. Accuracy does not move the same way. Individual rounds are
uneven, so a short window can dip below the cumulative average. On Banking20 the
uniform average is already within 0.5 points of its $T=20$ value by $T=5$ for
all three Qwen models, and on Newsgroups the curves are flat or mildly
non-monotone after $T=2$. The single best accuracy, however, often sits at an
intermediate horizon rather than at the endpoint ($T=18$ on 1.7B, $T=14$ on 4B,
and $T=18$ on 8B for Banking20). That is why $T=20$ uniform is the default: it
is the stable calibration rule, not the accuracy maximizer. The
accuracy-maximizing horizon is chosen after seeing the test set. The log-average
matches uniform accuracy at its own best horizon but remains less calibrated
(Table~\ref{tab:metrics}); its best NLL is $T=15$ on Banking20 with the 4B model
and $T=10$ on Newsgroups with the 1.7B model, rather than $T=20$.

\subsection{Runtime}
\label{sec:runtime}

Table~\ref{tab:runtime} reports seconds per decision on one RTX 4090 D, wall
time divided by the number of states, including batching and the pointwise pass.
Listwise is one forward pass per state. AnyJev-L0 performs $K$ cyclic reads and
reuses them for the batch prior. \method{} adds the pointwise reads and $TK$
listwise reads, and the tabulated time includes every round up to that $T$.
At $T=20$ the iterative readout costs $20$--$22\times$ AnyJev-L0, in line with
$TK$ versus $K$ reads; the pointwise pass is only 2--4\% of the total.
Newsgroups is slower than Banking20 because its prompts are longer. The method
is a compute-for-calibration trade-off. Section~\ref{sec:rounds} discusses    the practical
consequence: five to ten rounds already capture most of the NLL gain when a
full budget of 20 is too expensive.

\section{Conclusion}
\label{sec:conclusion}
To improve a System One choice decision, this paper introduces \method{},
which iterates that reorder-and-average step and
returns an average of the round distributions, so the decision does not rest on
the last round of a trajectory that, in our runs, has not converged.
On five benchmarks and five models the procedure has a flip rate of zero everywhere. Experiments also show \method{}'s advantage in calibration and accuracy.


\section*{Limitations}

The iteration costs about $T$ times one cyclic average. The round average is an
ensemble over the classes the trajectory happens to visit, not a proof of full
permutation symmetry, and Proposition~\ref{prop:termination} only guarantees
that a cycle exists somewhere in a state space too large to enumerate. The batch
prior assumes a representative evaluation batch; a pathological batch can
distort it, and items inside the batch are coupled through that prior. The
invariance proof assumes a deterministic model and canonically sorted prompts.
An interface that returns nondeterministic logits can still flip. The method
also inherits the pointwise initialization: when $\pi$ is nearly flat, the
initial order carries little information and later rounds can lose accuracy.

\bibliography{invariantjev}

\appendix

\section{Proofs}
\label{app:proofs}

\paragraph{Proof of Proposition~\ref{prop:invariance}.}
The initial order $\tau_1$ is invariant by Definition~\ref{def:invariant}. Assume
$\tau_t$ is invariant for every state. Every cyclic shift $\lambda_{t,s}$ is a
function of $\tau_t$. With a fixed prior, $b_{t,s}$ does not depend on the data
order. With the batch prior, $b_{t,s}$ is the mean over a canonically sorted
batch, or over content-free probes with the same listing, so it is a function of
the current invariant orders. Equation~\eqref{eq:round} is therefore a function
of $(x,q,\mathcal{O})$ and is invariant. Equation~\eqref{eq:reorder} reorders
that distribution and breaks ties by option text, so $\tau_{t+1}$ is invariant.
By induction every iterate is invariant. Any convex combination of invariant
distributions, including the uniform and exponential averages, is invariant.
The log-average is a deterministic function of the same distributions and is
invariant as well. \hfill$\square$

\paragraph{Proof of Proposition~\ref{prop:termination}.}
With a fixed prior, the transition is a deterministic map on a finite set of
$K!$ orders. Among any $K!+1$ consecutive orders, one repeats, so the sequence
enters a cycle. A cycle of period one is a fixed point.
With the batch prior, the prior at round $t$ depends on every state's order, so
the deterministic state is the joint vector of $N$ orders. There are $(K!)^N$
such vectors, and the same counting argument yields a cycle within
$(K!)^N+1$ rounds. Recomputing cyclic shifts from the current order does not
enlarge that set. \hfill$\square$

\section{Implementation details}
\label{app:implementation}

\paragraph{Prompts and readouts.}
The listwise prompt renders the question and the $K$ options under the letters
A, B, \dots, and we read the next-token probabilities of those letters at the
first answer position, renormalizing them over the $K$ letters to obtain
$g_\lambda$. The pointwise prompt shows one candidate and asks whether it is the
correct answer; we read the yes/no next-token log-odds
$f(o)=\log P(\mathrm{yes})-\log P(\mathrm{no})$ and set
$\pi=\mathrm{softmax}_o f(o)$. All prompts in a batch are rendered with the
model's chat template and sorted by prompt text before the forward pass, so the
batch composition and padding do not depend on the option order; ties in every
sort are broken by option text. Runs use fp16 with one forward pass per prompt
(only the first answer token is scored), batch sizes of 2--64 chosen per model
and benchmark to fit the 24\,GB device, seed 0 for the data order, and shuffle
seed 1234 for the flip measurements. The iteration is deterministic on a fixed
model and prompt set; the iterative flip-0 result follows from
Proposition~\ref{prop:invariance}, while the Listwise and AnyJev flips are
measured under one per-item shuffle.

\paragraph{Cost per decision.}
A listwise pass creates $K$ prompts per state, one per cyclic shift, and scores
the $K$ letters in each; the pointwise pass creates $K$ prompts. AnyJev-L0
therefore costs $K$ prompts per state and \method{} costs $K(T+1)$: the measured
$20$--$22\times$ ratio at $T=20$ is consistent with $TK$ versus $K$, and the
pointwise pass contributes only 2--4\% of the total.

\paragraph{Label priors.}
Both prior variants are distributions over list \emph{positions}, not over option
identities, and neither uses labels. For a listing $\lambda$ with $K$ positions,
the prior is $b_\lambda(j)=\frac{1}{|R|}\sum_{r\in R}g_\lambda^{(r)}(j)$, the mean
probability mass that the listwise readout places on position $j$ over a
label-free reference set $R$.
\begin{itemize}
\item \emph{Batch prior} (used in every experiment): $R$ is the evaluation batch
itself, that is, the states evaluated together, rather than a running estimate. It
is re-estimated for every cyclic shift $\lambda_{t,s}$ of the current order and
for every round, from the same listings that produce $g$, so it adds no forward
passes.
\item \emph{Per-item benchmarks}: option sets differ across questions, so states
are grouped by (question group, $K$, shift) and a separate prior is estimated for
each group with at least 8 states; smaller groups fall back to a prior pooled
over all states with the same $K$, and a shift with fewer than 2 states receives
no correction. In MMLU-Pro each of the 14 categories mixes several option counts,
so only 22 of its 83 (group, $K$) cells reach 8 states and the other 61 use the
$K$-pooled prior; Typed Decisions and JevBench have at most one option count per
group and use group priors throughout.
\item \emph{Content-free prior} (available, not used): $R$ consists of probes
with an empty state and the same question and listing. All reported AnyJev-L0
and \method{} numbers use the batch prior.
\end{itemize}

For each shift the prior is applied as
$p'_\lambda(j)\propto g_\lambda(j)/b_\lambda(j)^{\beta}$ with $\beta=1.0$,
evaluated in float64 with probabilities and priors clipped at
$\varepsilon=10^{-12}$ and renormalized; $\beta=0$ recovers the plain cyclic
average: S
\begin{equation}
p^{\perm}_{\sigma}(o)\;\propto\;
\exp\!\left(\frac{1}{K}\sum_{s=1}^{K}\log g_{\sigma_s}(o)\right).
\label{eq:perm}
\end{equation}
The corrected distributions enter the log-space
geometric mean over shifts in Eq.~\eqref{eq:round}. Because the prior is
recomputed from the current listings, it remains a function of the batch and of
the (invariant) current order; no labels or held-out data enter at any step.

\begin{table}[t]
\centering
\footnotesize
\setlength{\tabcolsep}{3pt}
\begin{tabular}{l p{4.5cm}}
\toprule
Setting & Value \\
\midrule
Prior type & batch position prior \\
Reference set & evaluation batch \\
Per-item grouping & (group, $K$, shift), $\ge 8$ states \\
Fallback & $K$-pooled; none if $<2$ states \\
Content-free probes & available, not used \\
Prior strength $\beta$ & 1.0 (AnyJev-L0 and \method{}) \\
Clipping $\varepsilon$ & $10^{-12}$ (float64) \\
Extra forward passes & none \\
Iterative horizon & $T{=}20$; best over $T{=}1,\dots,20$ \\
Data / shuffle seed & 0 / 1234 \\
Precision & fp16, one pass per prompt \\
\bottomrule
\end{tabular}
\caption{Implementation and label-prior settings.}
\label{tab:impl}
\end{table}

\section{Metrics}
 \label{app:metrics}
Let $p_i$ be the predicted distribution for item $i$, $y_i$ its label,
$\hat y_i=\arg\max_k p_i(k)$, and $\hat y_i'$ the prediction after one
independent per-item shuffle of the options. We report
\begin{align}\small
\mathrm{Acc} &= \frac{1}{N}\sum_{i=1}^{N}\mathbf{1}[\hat y_i=y_i],
\label{eq:acc}\\
\mathrm{Flip} &= \frac{1}{N}\sum_{i=1}^{N}\mathbf{1}[\hat y_i\neq\hat y_i'],
\label{eq:flip}\\
\mathrm{ECE} &= \sum_{b=1}^{B}\frac{|S_b|}{N}
  \bigl|\mathrm{acc}(S_b)-\mathrm{conf}(S_b)\bigr|,
\label{eq:ece}\\
\mathrm{Brier} &= \frac{1}{N}\sum_{i=1}^{N}\sum_{k=1}^{K_i}
  \bigl(p_i(k)-\mathbf{1}[y_i=k]\bigr)^2,
\label{eq:brier}\\
\mathrm{NLL} &= -\frac{1}{N}\sum_{i=1}^{N}\log p_i(y_i),
\label{eq:nll}
\end{align}
where $K_i$ is the number of options of item $i$. Accuracy ($\mathrm{Acc}$) is the fraction of
items whose top-1 option is correct and measures decision quality on the
original listing. $\mathrm{Flip}$ is the fraction whose top-1 option changes under one
independent shuffle; it measures input-order sensitivity directly, and an
order-invariant readout has flip 0. $\mathrm{ECE}$ compares confidence with accuracy over
$B=15$ equal-mass bins $S_b$: $\mathrm{conf}(S_b)$ is the mean top-1 probability
in the bin and $\mathrm{acc}(S_b)$ the empirical accuracy there, so ECE is the
bin-size-weighted average calibration gap (0 is perfect calibration). $\mathrm{Brier}$ is
the mean squared error of the whole predicted distribution over the $K_i$
options, a proper score that rewards distributions that are both calibrated and
sharp. $\mathrm{NLL}$ is the mean negative log-probability of the correct option, also a
proper score; it penalizes confident mistakes most heavily and is unbounded
above. Accuracy and flip rate are percentages; lower ECE, Brier, and NLL are better.
\section{Additional full tables}
\label{app:full-tables}

Tables~\ref{tab:main-full} and~\ref{tab:news-full} list every averaging
rule at $T=20$ for Banking20 and Newsgroups;
Tables~\ref{tab:rounds-bank} and~\ref{tab:rounds-news} give the
uniform and log-average rules at six selected horizons. The compact
versions in the main text are Tables~\ref{tab:all} and~\ref{tab:metrics}.

\begin{table*}[t]
\centering
\footnotesize
\setlength{\tabcolsep}{4pt}
\begin{tabular}{llccccc}
\toprule
Model & Method & Acc & Flip & ECE & Brier & NLL \\
\midrule
\multicolumn{7}{l}{\textit{Qwen3-1.7B, Banking20, full 799}} \\
 & Listwise (raw) & 62.1 & 33.7 & 0.355 & 0.728 & 6.560 \\
 & AnyJev-L0 & 69.5 & 11.8 & 0.270 & 0.571 & 3.382 \\
 & \method{} (last round) & 69.1 & \textbf{0.0} & 0.271 & 0.568 & 2.979 \\
 & \method{} (uniform average) & 70.3 & \textbf{0.0} & 0.190 & 0.471 & \textbf{1.839} \\
 & \method{} (log-average) & 70.5 & \textbf{0.0} & 0.254 & 0.532 & 2.619 \\
 & \method{} (exp.\ average, $\tau_w=10$) & 71.0 & \textbf{0.0} & 0.186 & 0.471 & 1.864 \\
 & \method{} (exp.\ average, $\tau_w=20$) & 70.8 & \textbf{0.0} & 0.186 & 0.470 & 1.848 \\
 & Best accuracy (post hoc, avg.\ last 3, $T=18$) & \textbf{72.0} & \textbf{0.0} & 0.193 & 0.487 & 2.276 \\
\midrule
\multicolumn{7}{l}{\textit{Qwen3-4B, Banking20, full 799}} \\
 & Listwise (raw) & 73.1 & 21.9 & 0.257 & 0.524 & 5.073 \\
 & AnyJev-L0 & 76.3 & 13.0 & 0.214 & 0.449 & 2.988 \\
 & \method{} (last round) & 75.5 & \textbf{0.0} & 0.226 & 0.468 & 3.102 \\
 & \method{} (uniform average) & 77.1 & \textbf{0.0} & 0.156 & 0.407 & \textbf{2.064} \\
 & \method{} (log-average) & 77.2 & \textbf{0.0} & 0.203 & 0.435 & 2.880 \\
 & \method{} (exp.\ average, $\tau_w=10$) & 76.6 & \textbf{0.0} & 0.164 & 0.411 & 2.074 \\
 & \method{} (exp.\ average, $\tau_w=20$) & 77.1 & \textbf{0.0} & 0.160 & 0.409 & 2.065 \\
 & Best accuracy (post hoc, uniform, $T=14$) & \textbf{77.6} & \textbf{0.0} & 0.157 & 0.406 & 2.155 \\
\midrule
\multicolumn{7}{l}{\textit{Qwen3-8B, Banking20, full 799}} \\
 & Listwise (raw) & 75.8 & 18.4 & 0.226 & 0.467 & 4.300 \\
 & AnyJev-L0 & 81.2 & 7.6 & 0.175 & 0.362 & 2.484 \\
 & \method{} (last round) & 79.7 & \textbf{0.0} & 0.186 & 0.378 & 2.630 \\
 & \method{} (uniform average) & 81.7 & \textbf{0.0} & 0.135 & 0.332 & \textbf{1.685} \\
 & \method{} (log-average) & 81.9 & \textbf{0.0} & 0.166 & 0.350 & 2.425 \\
 & \method{} (exp.\ average, $\tau_w=10$) & 81.6 & \textbf{0.0} & 0.135 & 0.335 & 1.707 \\
 & \method{} (exp.\ average, $\tau_w=20$) & 81.7 & \textbf{0.0} & 0.136 & 0.333 & 1.693 \\
 & Best accuracy (post hoc, uniform, $T=18$) & \textbf{82.0} & \textbf{0.0} & 0.133 & 0.332 & 1.697 \\
\bottomrule
\end{tabular}
\caption{Banking20 full-pool results (799 items). Acc and Flip are percentages;
ECE, Brier, and NLL are lower-is-better. AnyJev-L0 uses the library's full batch
prior; iterative runs use $\beta=1.0$ and $T=20$.
Iterative flip is 0.0 by Proposition~\ref{prop:invariance}; the 20-round runs were
executed on the original order only, while the AnyJev flip rates are
measured under the standard shuffle. The last round is shown as a control and is
not the recommended rule. The uniform $T=20$ row has the best NLL; the
best-accuracy rows select the horizon on the same test set and are descriptive.}
\label{tab:main-full}
\end{table*}

\begin{table*}[t]
\centering
\footnotesize
\setlength{\tabcolsep}{4pt}
\begin{tabular}{lccccc}
\toprule
Method & Acc & Flip & ECE & Brier & NLL \\
\midrule
\multicolumn{6}{l}{\textit{Qwen3-1.7B, Newsgroups, 300}} \\
Listwise (raw) & 59.0 & 28.7 & 0.387 & 0.790 & 7.026 \\
AnyJev-L0 & 61.3 & 19.7 & 0.349 & 0.720 & 4.719 \\
\method{} (last round) & 60.0 & 0.0 & 0.358 & 0.732 & 4.484 \\
\method{} (uniform average) & 62.0 & 0.0 & 0.275 & 0.635 & \textbf{2.983} \\
\method{} (log-average) & 63.0 & 0.0 & 0.324 & 0.689 & 4.124 \\
\method{} (exp.\ average, $\tau_w=10$) & 62.3 & 0.0 & 0.279 & 0.644 & 3.021 \\
\method{} (exp.\ average, $\tau_w=20$) & 62.3 & 0.0 & 0.276 & 0.640 & 2.997 \\
Best accuracy (post hoc, logavg, $T=2$) & \textbf{65.3} & 0.0 & 0.312 & 0.661 & 4.108 \\
\midrule
\multicolumn{6}{l}{\textit{Qwen3-4B, Newsgroups, 300}} \\
Listwise (raw) & 63.7 & 20.0 & 0.340 & 0.692 & 6.171 \\
AnyJev-L0 & 66.7 & 11.7 & 0.310 & 0.631 & 4.717 \\
\method{} (last round) & 66.3 & 0.0 & 0.302 & 0.630 & 4.126 \\
\method{} (uniform average) & 67.3 & 0.0 & 0.238 & 0.549 & \textbf{2.702} \\
\method{} (log-average) & 67.7 & 0.0 & 0.298 & 0.611 & 3.770 \\
\method{} (exp.\ average, $\tau_w=10$) & 68.0 & 0.0 & 0.235 & 0.550 & 2.741 \\
\method{} (exp.\ average, $\tau_w=20$) & 68.0 & 0.0 & 0.234 & 0.549 & 2.718 \\
Best accuracy (post hoc, last, $T=19$) & \textbf{69.7} & 0.0 & 0.277 & 0.591 & 4.139 \\
\midrule
\multicolumn{6}{l}{\textit{Qwen3-8B, Newsgroups, 300}} \\
Listwise (raw) & 64.3 & 20.7 & 0.331 & 0.680 & 5.725 \\
AnyJev-L0 & 66.3 & 12.7 & 0.306 & 0.630 & 4.268 \\
\method{} (last round) & 66.0 & 0.0 & 0.310 & 0.650 & 4.325 \\
\method{} (uniform average) & 66.0 & 0.0 & 0.256 & 0.581 & \textbf{2.919} \\
\method{} (log-average) & 65.3 & 0.0 & 0.316 & 0.635 & 4.093 \\
\method{} (exp.\ average, $\tau_w=10$) & 65.7 & 0.0 & 0.263 & 0.584 & 2.938 \\
\method{} (exp.\ average, $\tau_w=20$) & 65.7 & 0.0 & 0.261 & 0.583 & 2.923 \\
Best accuracy (post hoc, uniform, $T=9$) & \textbf{66.7} & 0.0 & 0.246 & 0.583 & 3.103 \\
\bottomrule
\end{tabular}
\caption{Newsgroups, 300 items. Fixed rules use $T=20$; Acc and Flip are
percentages. Iterative flip is 0 by Proposition~\ref{prop:invariance}; the
iterative runs were executed on the original order only. The uniform average's
Brier/NLL improvements over AnyJev-L0 are significant on 1.7B/4B under the
paired bootstrap; on 8B the NLL gain is significant and the Brier gain is
borderline. Accuracy gains under the fixed rule (1--2 items) are not
significant. The best-accuracy rows select the horizon on the same test set and
are descriptive.}
\label{tab:news-full}
\end{table*}

\begin{table*}[t]
\centering
\footnotesize
\setlength{\tabcolsep}{5pt}
\begin{tabular}{llccc}
\toprule
Model & $T$ & Uniform avg Acc / ECE / NLL & Log-average Acc / ECE / NLL & s/dec \\
\midrule
1.7B & 1 & 69.5 / 0.262 / 3.128 & 69.5 / 0.262 / 3.128 & 0.211 \\
     & 2 & 70.5 / 0.205 / 2.427 & 70.8 / 0.249 / 2.779 & 0.363 \\
     & 5 & 70.6 / 0.193 / 2.107 & 70.7 / 0.249 / 2.690 & 0.819 \\
     & 10 & \textbf{70.8} / 0.189 / 1.951 & \textbf{70.8} / 0.249 / 2.678 & 1.578 \\
     & 15 & 70.2 / 0.194 / 1.911 & 70.1 / 0.257 / 2.668 & 2.337 \\
     & 20 & 70.3 / 0.190 / \textbf{1.839} & 70.5 / 0.254 / \textbf{2.619} & 3.096 \\
\midrule
4B   & 1 & 76.0 / 0.223 / 3.611 & 76.0 / 0.223 / 3.611 & 0.483 \\
     & 2 & 75.8 / 0.179 / 2.705 & 76.2 / 0.218 / 3.098 & 0.825 \\
     & 5 & 76.8 / 0.168 / 2.426 & 77.0 / 0.211 / 2.994 & 1.850 \\
     & 10 & 76.2 / 0.165 / 2.206 & 76.6 / 0.212 / 2.887 & 3.557 \\
     & 15 & \textbf{77.3} / 0.153 / 2.141 & \textbf{77.2} / 0.203 / \textbf{2.873} & 5.265 \\
     & 20 & 77.1 / 0.156 / \textbf{2.064} & 77.2 / 0.203 / 2.880 & 6.972 \\
\midrule
8B   & 1 & 80.4 / 0.181 / 2.908 & 80.4 / 0.181 / 2.908 & 0.821 \\
     & 2 & 81.4 / 0.149 / 2.276 & 81.6 / 0.170 / 2.588 & 1.413 \\
     & 5 & \textbf{82.0} / 0.135 / 2.014 & 81.7 / 0.168 / 2.512 & 3.193 \\
     & 10 & 81.4 / 0.132 / 1.795 & 81.5 / 0.168 / 2.427 & 6.146 \\
     & 15 & 81.5 / 0.131 / 1.751 & 81.5 / 0.168 / 2.432 & 9.096 \\
     & 20 & 81.7 / 0.135 / \textbf{1.685} & \textbf{81.9} / 0.166 / \textbf{2.425} & 12.045 \\
\bottomrule
\end{tabular}
\caption{Round-count ablation for Banking20: uniform average versus log-average
at $T=1,2,5,10,15,20$. $T=1$ is a single cyclic average from the pointwise
order. Accuracy in percent; ECE and NLL lower-is-better; s/dec is seconds per
decision including the pointwise pass. Bold marks the best accuracy for each
rule and model among these horizons; bold NLL marks the best NLL. The full
$T=1,\dots,20$ tables are in Appendix~\ref{app:rounds-full}.}
\label{tab:rounds-bank}
\end{table*}

\begin{table*}[t]
\centering
\footnotesize
\setlength{\tabcolsep}{5pt}
\begin{tabular}{llccc}
\toprule
Model & $T$ & Uniform avg Acc / ECE / NLL & Log-average Acc / ECE / NLL & s/dec \\
\midrule
1.7B & 1 & \textbf{65.0} / 0.317 / 4.795 & 65.0 / 0.317 / 4.795 & 0.445 \\
     & 2 & 64.3 / 0.257 / 3.635 & \textbf{65.3} / 0.312 / 4.108 & 0.697 \\
     & 5 & 63.0 / 0.267 / 3.254 & 63.0 / 0.328 / 4.105 & 1.454 \\
     & 10 & 61.3 / 0.278 / 3.083 & 62.7 / 0.323 / \textbf{4.083} & 2.715 \\
     & 15 & 61.7 / 0.277 / 3.045 & 62.0 / 0.332 / 4.118 & 3.978 \\
     & 20 & 62.0 / 0.275 / \textbf{2.983} & 63.0 / 0.324 / 4.124 & 5.239 \\
\midrule
4B   & 1 & 65.7 / 0.297 / 4.540 & 65.7 / 0.297 / 4.540 & 1.003 \\
     & 2 & \textbf{68.0} / 0.233 / 3.434 & 67.7 / 0.291 / 3.867 & 1.571 \\
     & 5 & 67.0 / 0.242 / 2.974 & 67.7 / 0.293 / 3.849 & 3.273 \\
     & 10 & 66.3 / 0.245 / 2.823 & 66.7 / 0.306 / 3.825 & 6.110 \\
     & 15 & 67.3 / 0.238 / 2.702 & 67.3 / 0.302 / 3.793 & 8.947 \\
     & 20 & 67.3 / 0.238 / \textbf{2.702} & \textbf{67.7} / 0.298 / \textbf{3.770} & 11.785 \\
\midrule
8B   & 1 & 64.7 / 0.318 / 4.553 & 64.7 / 0.318 / 4.553 & 1.617 \\
     & 2 & 65.0 / 0.261 / 3.626 & 65.3 / 0.311 / 4.150 & 2.534 \\
     & 5 & 65.7 / 0.254 / 3.286 & \textbf{66.3} / 0.303 / 4.129 & 5.284 \\
     & 10 & \textbf{66.3} / 0.251 / 3.084 & 66.0 / 0.308 / 4.116 & 9.869 \\
     & 15 & 65.7 / 0.259 / 3.014 & 65.7 / 0.312 / 4.118 & 14.455 \\
     & 20 & 66.0 / 0.256 / \textbf{2.919} & 65.3 / 0.316 / \textbf{4.093} & 19.041 \\
\bottomrule
\end{tabular}
\caption{Newsgroups, 300 items: uniform average versus log-average at
$T=1,2,5,10,15,20$. Bold marks the best accuracy for each rule and model among
these horizons and the best NLL.}
\label{tab:rounds-news}
\end{table*}

\section{Per-round results}
\label{app:rounds-full}

Tables~\ref{tab:rounds-17}, \ref{tab:rounds-4}, \ref{tab:rounds-8B},
\ref{tab:rounds-news-17}, \ref{tab:rounds-news-4} and~\ref{tab:rounds-news-8B}
record the average over rounds $1,\dots,T$ for every horizon. ``Best at $T$'' is the rule with the
highest accuracy at that horizon; the value after the slash is its NLL. The last
column is seconds per decision including the pointwise pass.

\begin{table*}[h]
\centering
\footnotesize
\setlength{\tabcolsep}{3.5pt}
\begin{tabular}{clccc}
\toprule
$T$ & Uniform Acc / ECE / NLL & Exp.\ $\tau_w=10$ Acc / ECE / NLL & Best at $T$ & s/dec \\
\midrule
1 & 69.5 / 0.262 / 3.128 & 69.5 / 0.262 / 3.128 & uniform 69.5 / 3.128 & 0.211 \\
2 & 70.5 / 0.205 / 2.427 & 70.1 / 0.209 / 2.426 & logavg 70.8 / 2.779 & 0.363 \\
3 & 69.8 / 0.213 / 2.274 & 69.8 / 0.214 / 2.275 & logavg 70.7 / 2.778 & 0.515 \\
4 & 71.1 / 0.187 / 2.143 & 71.2 / 0.185 / 2.143 & exp10 71.2 / 2.143 & 0.667 \\
5 & 70.6 / 0.193 / 2.107 & 70.3 / 0.195 / 2.110 & logavg 70.7 / 2.690 & 0.819 \\
6 & 70.8 / 0.188 / 2.064 & 70.8 / 0.188 / 2.068 & avg\_last3 71.1 / 2.324 & 0.971 \\
7 & 70.5 / 0.193 / 2.011 & 70.3 / 0.195 / 2.016 & logavg 71.0 / 2.675 & 1.122 \\
8 & 70.8 / 0.188 / 1.994 & 71.0 / 0.188 / 2.003 & exp20 71.0 / 1.997 & 1.274 \\
9 & 71.0 / 0.186 / 1.960 & 70.3 / 0.193 / 1.972 & uniform 71.0 / 1.960 & 1.426 \\
10 & 70.8 / 0.189 / 1.951 & 70.7 / 0.191 / 1.964 & uniform 70.8 / 1.951 & 1.578 \\
11 & 70.6 / 0.191 / 1.958 & 70.3 / 0.195 / 1.978 & uniform 70.6 / 1.958 & 1.729 \\
12 & 70.5 / 0.191 / 1.963 & 70.2 / 0.194 / 1.990 & uniform 70.5 / 1.963 & 1.881 \\
13 & 70.5 / 0.191 / 1.944 & 70.2 / 0.196 / 1.968 & avg\_last3 70.7 / 2.506 & 2.033 \\
14 & 70.3 / 0.192 / 1.926 & 70.5 / 0.192 / 1.948 & exp10 70.5 / 1.948 & 2.185 \\
15 & 70.2 / 0.194 / 1.911 & 70.5 / 0.194 / 1.937 & avg\_last3 71.2 / 2.344 & 2.337 \\
16 & 70.2 / 0.194 / 1.903 & 70.6 / 0.192 / 1.930 & avg\_last5 71.3 / 2.221 & 2.489 \\
17 & 70.3 / 0.193 / 1.891 & 70.5 / 0.195 / 1.921 & avg\_last5 71.2 / 2.196 & 2.640 \\
18 & 70.6 / 0.189 / 1.872 & 70.6 / 0.190 / 1.901 & avg\_last3 72.0 / 2.276 & 2.792 \\
19 & 70.3 / 0.191 / 1.861 & 70.8 / 0.188 / 1.888 & avg\_last3 71.2 / 2.285 & 2.944 \\
20 & 70.3 / 0.190 / 1.839 & 71.0 / 0.186 / 1.864 & avg\_last3 71.1 / 2.223 & 3.096 \\
\bottomrule
\end{tabular}
\caption{Qwen3-1.7B, Banking20, full 799, every horizon.}
\label{tab:rounds-17}
\end{table*}

\begin{table*}[h]
\centering
\footnotesize
\setlength{\tabcolsep}{3.5pt}
\begin{tabular}{clccc}
\toprule
$T$ & Uniform Acc / ECE / NLL & Exp.\ $\tau_w=10$ Acc / ECE / NLL & Best at $T$ & s/dec \\
\midrule
1 & 76.0 / 0.223 / 3.611 & 76.0 / 0.223 / 3.611 & uniform 76.0 / 3.611 & 0.483 \\
2 & 75.8 / 0.179 / 2.705 & 74.7 / 0.190 / 2.700 & logavg 76.2 / 3.098 & 0.825 \\
3 & 76.0 / 0.182 / 2.524 & 76.1 / 0.181 / 2.521 & logavg 76.8 / 3.043 & 1.166 \\
4 & 76.3 / 0.170 / 2.437 & 76.0 / 0.173 / 2.436 & logavg 76.8 / 2.972 & 1.508 \\
5 & 76.8 / 0.168 / 2.426 & 77.1 / 0.174 / 2.430 & exp10 77.1 / 2.430 & 1.850 \\
6 & 76.8 / 0.166 / 2.345 & 76.6 / 0.163 / 2.346 & logavg 77.1 / 2.931 & 2.191 \\
7 & 77.2 / 0.164 / 2.319 & 77.0 / 0.162 / 2.322 & uniform 77.2 / 2.319 & 2.533 \\
8 & 76.6 / 0.162 / 2.262 & 76.7 / 0.161 / 2.264 & exp20 76.8 / 2.262 & 2.874 \\
9 & 77.0 / 0.167 / 2.257 & 77.1 / 0.163 / 2.263 & exp10 77.1 / 2.263 & 3.216 \\
10 & 76.2 / 0.165 / 2.206 & 76.2 / 0.166 / 2.210 & logavg 76.6 / 2.887 & 3.557 \\
11 & 76.6 / 0.161 / 2.202 & 76.7 / 0.162 / 2.207 & exp20 76.7 / 2.203 & 3.899 \\
12 & 77.0 / 0.157 / 2.186 & 76.8 / 0.159 / 2.194 & logavg 77.1 / 2.878 & 4.240 \\
13 & 77.5 / 0.157 / 2.173 & 77.0 / 0.158 / 2.181 & exp20 77.6 / 2.175 & 4.582 \\
14 & 77.6 / 0.157 / 2.155 & 77.0 / 0.159 / 2.163 & uniform 77.6 / 2.155 & 4.924 \\
15 & 77.3 / 0.153 / 2.141 & 76.6 / 0.163 / 2.148 & uniform 77.3 / 2.141 & 5.265 \\
16 & 77.0 / 0.157 / 2.127 & 76.7 / 0.164 / 2.136 & uniform 77.0 / 2.127 & 5.607 \\
17 & 76.6 / 0.161 / 2.120 & 76.6 / 0.165 / 2.133 & logavg 77.0 / 2.882 & 5.948 \\
18 & 76.8 / 0.158 / 2.106 & 76.8 / 0.163 / 2.121 & uniform 76.8 / 2.106 & 6.290 \\
19 & 77.0 / 0.157 / 2.092 & 77.0 / 0.163 / 2.108 & exp20 77.1 / 2.096 & 6.631 \\
20 & 77.1 / 0.156 / 2.064 & 76.6 / 0.164 / 2.074 & logavg 77.2 / 2.880 & 6.972 \\
\bottomrule
\end{tabular}
\caption{Qwen3-4B, Banking20, full 799, every horizon.}
\label{tab:rounds-4}
\end{table*}

\begin{table*}[h]
\centering
\footnotesize
\setlength{\tabcolsep}{3.5pt}
\begin{tabular}{clccc}
\toprule
$T$ & Uniform Acc / ECE / NLL & Exp.\ $\tau_w=10$ Acc / ECE / NLL & Best at $T$ & s/dec \\
\midrule
1 & 80.4 / 0.181 / 2.908 & 80.4 / 0.181 / 2.908 & uniform 80.4 / 2.908 & 0.821 \\
2 & 81.4 / 0.149 / 2.276 & 81.4 / 0.149 / 2.273 & logavg 81.6 / 2.588 & 1.413 \\
3 & 81.2 / 0.146 / 2.170 & 81.1 / 0.147 / 2.167 & logavg 81.7 / 2.579 & 2.005 \\
4 & 81.6 / 0.137 / 2.056 & 81.9 / 0.135 / 2.051 & exp10 81.9 / 2.051 & 2.598 \\
5 & 82.0 / 0.135 / 2.014 & 81.6 / 0.136 / 2.012 & uniform 82.0 / 2.014 & 3.193 \\
6 & 82.0 / 0.134 / 1.937 & 81.5 / 0.135 / 1.934 & uniform 82.0 / 1.937 & 3.786 \\
7 & 81.6 / 0.133 / 1.919 & 81.2 / 0.137 / 1.919 & logavg 82.0 / 2.475 & 4.377 \\
8 & 81.6 / 0.135 / 1.857 & 81.4 / 0.135 / 1.858 & exp20 81.7 / 1.857 & 4.967 \\
9 & 81.6 / 0.132 / 1.826 & 81.1 / 0.136 / 1.829 & logavg 81.9 / 2.434 & 5.556 \\
10 & 81.4 / 0.132 / 1.795 & 80.9 / 0.137 / 1.796 & logavg 81.5 / 2.427 & 6.146 \\
11 & 81.5 / 0.131 / 1.802 & 81.1 / 0.138 / 1.810 & uniform 81.5 / 1.802 & 6.736 \\
12 & 81.5 / 0.135 / 1.783 & 81.5 / 0.133 / 1.791 & uniform 81.5 / 1.783 & 7.326 \\
13 & 81.7 / 0.133 / 1.777 & 80.9 / 0.139 / 1.787 & uniform 81.7 / 1.777 & 7.916 \\
14 & 81.4 / 0.132 / 1.754 & 81.0 / 0.137 / 1.763 & logavg 81.5 / 2.429 & 8.506 \\
15 & 81.5 / 0.131 / 1.751 & 81.0 / 0.137 / 1.761 & uniform 81.5 / 1.751 & 9.096 \\
16 & 81.5 / 0.133 / 1.744 & 81.2 / 0.134 / 1.758 & uniform 81.5 / 1.744 & 9.685 \\
17 & 81.9 / 0.132 / 1.730 & 81.5 / 0.135 / 1.748 & uniform 81.9 / 1.730 & 10.275 \\
18 & 82.0 / 0.133 / 1.697 & 81.4 / 0.133 / 1.716 & uniform 82.0 / 1.697 & 10.865 \\
19 & 81.7 / 0.130 / 1.696 & 81.2 / 0.137 / 1.718 & uniform 81.7 / 1.696 & 11.455 \\
20 & 81.7 / 0.135 / 1.685 & 81.6 / 0.135 / 1.707 & logavg 81.9 / 2.425 & 12.045 \\
\bottomrule
\end{tabular}
\caption{Qwen3-8B, Banking20, full 799, every horizon.}
\label{tab:rounds-8B}
\end{table*}

\begin{table*}[h]
\centering
\footnotesize
\setlength{\tabcolsep}{3.5pt}
\begin{tabular}{clccc}
\toprule
$T$ & Uniform Acc / ECE / NLL & Exp.\ $\tau_w=10$ Acc / ECE / NLL & Best at $T$ & s/dec \\
\midrule
1 & 65.0 / 0.317 / 4.795 & 65.0 / 0.317 / 4.795 & uniform 65.0 / 4.795 & 0.445 \\
2 & 64.3 / 0.257 / 3.635 & 62.3 / 0.277 / 3.632 & logavg 65.3 / 4.108 & 0.697 \\
3 & 62.3 / 0.277 / 3.418 & 62.3 / 0.278 / 3.415 & logavg 64.0 / 4.067 & 0.949 \\
4 & 62.7 / 0.268 / 3.319 & 62.3 / 0.272 / 3.319 & logavg 63.3 / 4.032 & 1.202 \\
5 & 63.0 / 0.267 / 3.254 & 63.0 / 0.269 / 3.258 & uniform 63.0 / 3.254 & 1.454 \\
6 & 62.7 / 0.265 / 3.198 & 62.3 / 0.269 / 3.205 & avg\_last5 63.3 / 3.294 & 1.706 \\
7 & 63.0 / 0.263 / 3.183 & 63.0 / 0.265 / 3.196 & exp20 63.3 / 3.189 & 1.958 \\
8 & 62.3 / 0.267 / 3.147 & 61.7 / 0.276 / 3.160 & uniform 62.3 / 3.147 & 2.211 \\
9 & 62.3 / 0.266 / 3.104 & 62.0 / 0.270 / 3.119 & logavg 63.0 / 4.073 & 2.463 \\
10 & 61.3 / 0.278 / 3.083 & 61.3 / 0.283 / 3.098 & logavg 62.7 / 4.083 & 2.715 \\
11 & 62.0 / 0.273 / 3.064 & 62.0 / 0.278 / 3.077 & logavg 62.3 / 4.088 & 2.968 \\
12 & 61.0 / 0.283 / 3.049 & 61.0 / 0.288 / 3.065 & avg\_last5 61.7 / 3.454 & 3.220 \\
13 & 61.7 / 0.276 / 3.043 & 62.0 / 0.277 / 3.065 & logavg 62.3 / 4.105 & 3.473 \\
14 & 61.3 / 0.279 / 3.038 & 61.0 / 0.288 / 3.065 & avg\_last3 61.7 / 3.843 & 3.725 \\
15 & 61.7 / 0.277 / 3.045 & 62.0 / 0.280 / 3.079 & exp10 62.0 / 3.079 & 3.978 \\
16 & 61.3 / 0.281 / 3.025 & 61.7 / 0.284 / 3.057 & avg\_last5 62.3 / 3.659 & 4.230 \\
17 & 62.0 / 0.274 / 3.032 & 62.7 / 0.275 / 3.074 & exp10 62.7 / 3.074 & 4.482 \\
18 & 62.0 / 0.275 / 3.015 & 62.7 / 0.275 / 3.057 & avg\_last3 63.3 / 3.778 & 4.734 \\
19 & 61.7 / 0.278 / 2.992 & 63.0 / 0.272 / 3.033 & exp10 63.0 / 3.033 & 4.987 \\
20 & 62.0 / 0.275 / 2.983 & 62.3 / 0.279 / 3.021 & avg\_last3 63.0 / 3.730 & 5.239 \\
\bottomrule
\end{tabular}
\caption{Qwen3-1.7B, Newsgroups, 300, every horizon.}
\label{tab:rounds-news-17}
\end{table*}

\begin{table*}[h]
\centering
\footnotesize
\setlength{\tabcolsep}{3.5pt}
\begin{tabular}{clccc}
\toprule
$T$ & Uniform Acc / ECE / NLL & Exp.\ $\tau_w=10$ Acc / ECE / NLL & Best at $T$ & s/dec \\
\midrule
1 & 65.7 / 0.297 / 4.540 & 65.7 / 0.297 / 4.540 & uniform 65.7 / 4.540 & 1.003 \\
2 & 68.0 / 0.233 / 3.434 & 69.0 / 0.224 / 3.431 & exp10 69.0 / 3.431 & 1.571 \\
3 & 67.0 / 0.247 / 3.162 & 67.0 / 0.249 / 3.158 & logavg 68.0 / 3.854 & 2.138 \\
4 & 68.7 / 0.222 / 3.066 & 68.7 / 0.222 / 3.064 & exp10 68.7 / 3.064 & 2.706 \\
5 & 67.0 / 0.242 / 2.974 & 66.7 / 0.246 / 2.973 & logavg 67.7 / 3.849 & 3.273 \\
6 & 67.3 / 0.232 / 2.890 & 67.3 / 0.232 / 2.891 & avg\_last5 68.0 / 2.920 & 3.840 \\
7 & 66.7 / 0.241 / 2.868 & 66.3 / 0.245 / 2.873 & logavg 68.0 / 3.809 & 4.408 \\
8 & 67.0 / 0.238 / 2.829 & 67.0 / 0.240 / 2.842 & avg\_last5 67.3 / 3.044 & 4.975 \\
9 & 66.3 / 0.247 / 2.833 & 66.3 / 0.250 / 2.852 & logavg 66.7 / 3.835 & 5.543 \\
10 & 66.3 / 0.245 / 2.823 & 66.7 / 0.245 / 2.841 & exp20 66.7 / 2.830 & 6.110 \\
11 & 66.3 / 0.248 / 2.810 & 66.7 / 0.248 / 2.827 & logavg 67.0 / 3.834 & 6.677 \\
12 & 67.0 / 0.241 / 2.730 & 67.3 / 0.241 / 2.743 & exp20 67.3 / 2.734 & 7.244 \\
13 & 66.7 / 0.245 / 2.726 & 67.0 / 0.246 / 2.742 & last 67.7 / 4.046 & 7.812 \\
14 & 67.3 / 0.238 / 2.716 & 67.7 / 0.239 / 2.737 & avg\_last5 68.0 / 2.978 & 8.379 \\
15 & 67.3 / 0.238 / 2.702 & 67.3 / 0.241 / 2.721 & avg\_last3 68.0 / 3.219 & 8.947 \\
16 & 67.3 / 0.239 / 2.708 & 67.7 / 0.240 / 2.737 & avg\_last5 68.3 / 3.019 & 9.514 \\
17 & 67.3 / 0.240 / 2.709 & 68.3 / 0.236 / 2.743 & exp10 68.3 / 2.743 & 10.082 \\
18 & 67.3 / 0.239 / 2.711 & 68.0 / 0.237 / 2.748 & avg\_last5 68.7 / 3.086 & 10.650 \\
19 & 67.3 / 0.239 / 2.709 & 68.3 / 0.234 / 2.746 & last 69.7 / 4.139 & 11.217 \\
20 & 67.3 / 0.238 / 2.702 & 68.0 / 0.235 / 2.741 & avg\_last3 68.3 / 3.300 & 11.785 \\
\bottomrule
\end{tabular}
\caption{Qwen3-4B, Newsgroups, 300, every horizon.}
\label{tab:rounds-news-4}
\end{table*}

\begin{table*}[h]
\centering
\footnotesize
\setlength{\tabcolsep}{3.5pt}
\begin{tabular}{clccc}
\toprule
$T$ & Uniform Acc / ECE / NLL & Exp.\ $\tau_w=10$ Acc / ECE / NLL & Best at $T$ & s/dec \\
\midrule
1 & 64.7 / 0.318 / 4.553 & 64.7 / 0.318 / 4.553 & uniform 64.7 / 4.553 & 1.617 \\
2 & 65.0 / 0.261 / 3.626 & 64.0 / 0.271 / 3.623 & logavg 65.3 / 4.150 & 2.534 \\
3 & 65.0 / 0.270 / 3.557 & 65.0 / 0.271 / 3.561 & logavg 65.3 / 4.242 & 3.450 \\
4 & 65.0 / 0.262 / 3.329 & 64.7 / 0.266 / 3.328 & logavg 65.3 / 4.123 & 4.367 \\
5 & 65.7 / 0.254 / 3.286 & 65.7 / 0.255 / 3.287 & avg\_last3 66.3 / 3.509 & 5.284 \\
6 & 65.7 / 0.250 / 3.250 & 66.0 / 0.247 / 3.254 & exp10 66.0 / 3.254 & 6.201 \\
7 & 66.0 / 0.252 / 3.197 & 66.3 / 0.252 / 3.199 & last 66.7 / 4.354 & 7.118 \\
8 & 66.3 / 0.247 / 3.104 & 66.0 / 0.253 / 3.103 & exp20 66.3 / 3.102 & 8.035 \\
9 & 66.7 / 0.246 / 3.103 & 66.7 / 0.249 / 3.111 & uniform 66.7 / 3.103 & 8.952 \\
10 & 66.3 / 0.251 / 3.084 & 66.3 / 0.255 / 3.096 & uniform 66.3 / 3.084 & 9.869 \\
11 & 66.3 / 0.253 / 3.049 & 66.0 / 0.261 / 3.057 & uniform 66.3 / 3.049 & 10.786 \\
12 & 66.3 / 0.252 / 3.023 & 65.7 / 0.261 / 3.031 & uniform 66.3 / 3.023 & 11.703 \\
13 & 66.3 / 0.252 / 3.024 & 65.7 / 0.262 / 3.038 & uniform 66.3 / 3.024 & 12.620 \\
14 & 66.0 / 0.255 / 3.018 & 65.3 / 0.265 / 3.039 & avg\_last3 66.7 / 3.606 & 13.537 \\
15 & 65.7 / 0.259 / 3.014 & 65.3 / 0.266 / 3.035 & last 66.0 / 4.422 & 14.455 \\
16 & 66.0 / 0.255 / 3.005 & 65.7 / 0.260 / 3.033 & avg\_last3 66.3 / 3.629 & 15.372 \\
17 & 66.0 / 0.255 / 2.981 & 66.3 / 0.254 / 3.005 & exp10 66.3 / 3.005 & 16.289 \\
18 & 66.0 / 0.254 / 2.965 & 65.3 / 0.262 / 2.990 & uniform 66.0 / 2.965 & 17.206 \\
19 & 65.7 / 0.259 / 2.943 & 65.7 / 0.262 / 2.967 & avg\_last5 66.7 / 3.355 & 18.123 \\
20 & 66.0 / 0.256 / 2.919 & 65.7 / 0.263 / 2.938 & uniform 66.0 / 2.919 & 19.041 \\
\bottomrule
\end{tabular}
\caption{Qwen3-8B, Newsgroups, 300, every horizon.}
\label{tab:rounds-news-8B}
\end{table*}

\end{document}